\documentclass[11pt,a4paper]{article}

\usepackage[margin=1in]{geometry}
\usepackage{libertinus}          
\usepackage{microtype}
\usepackage{parskip}
\usepackage{titlesec}
\titleformat{\section}
  {\normalfont\Large\bfseries}
  {\thesection}{0.75em}{}[\vspace{0.25ex}\titlerule]
\titleformat{\subsection}
  {\normalfont\large\bfseries}
  {\thesubsection}{0.6em}{}
\titleformat{\subsubsection}
  {\normalfont\normalsize\bfseries}
  {\thesubsubsection}{0.55em}{}
\titlespacing*{\section}{0pt}{2.2ex plus .8ex minus .2ex}{1.0ex}
\titlespacing*{\subsection}{0pt}{1.8ex plus .7ex minus .2ex}{0.8ex}
\titlespacing*{\subsubsection}{0pt}{1.4ex plus .5ex minus .2ex}{0.6ex}

\usepackage{enumitem}
\setlist{nosep,leftmargin=*}

\usepackage[hidelinks]{hyperref}
\usepackage{orcidlink}

\usepackage{comment} 
\usepackage{caption}
\usepackage{amsmath}
\usepackage{float}
\usepackage[utf8x]{inputenc}
\usepackage{amsmath,amssymb}
\usepackage{xcolor}
\usepackage{mathtools}
\usepackage{tabulary}
\usepackage{amsthm}
\usepackage{latexsym}
\usepackage[mathcal]{eucal}
\usepackage{amscd}
\usepackage{graphicx}
\usepackage{amsfonts}
\usepackage{amscd}
\usepackage{MnSymbol}
\usepackage{setspace}

\usepackage[english]{babel}
\usepackage{listings}
\usepackage{calligra}
\usepackage{titlesec}
\usepackage{booktabs}

\newtheorem{proposition}{Proposition}[section]

\theoremstyle{remark}

\title{\textbf{\Large  Echo State Networks as State-Space Models: A Systems Perspective}}

\author{
    \large 
    Pradeep Singh\orcidlink{0000-0002-5372-3355}\thanks{Email: \texttt{pradeep.cs@sric.iitr.ac.in}},     
    Balasubramanian Raman\orcidlink{0000-0001-6277-6267}\thanks{Email: \texttt{bala@cs.iitr.ac.in}}\vspace{0.2cm}\\
    \begin{minipage}[t]{0.5\textwidth}
    \centering
    \small Department of Computer Science and Engineering\\
    \small Indian Institute of Technology Roorkee\\
    \small Roorkee-247667, India
    \end{minipage}
}

\date{}

\begin{document}\maketitle

\begin{abstract}
Echo State Networks (ESNs) are typically presented as efficient, readout-trained recurrent models, yet their dynamics and design are often guided by heuristics rather than first principles. We recast ESNs explicitly as state-space models (SSMs), providing a unified systems-theoretic account that links reservoir computing with classical identification and modern kernelized SSMs. First, we show that the echo-state property is an instance of input-to-state stability for a contractive nonlinear SSM and derive verifiable conditions in terms of leak, spectral scaling, and activation Lipschitz constants. Second, we develop two complementary mappings: (i) small-signal linearizations that yield locally valid LTI SSMs with interpretable poles and memory horizons; and (ii) lifted/Koopman random-feature expansions that render the ESN a linear SSM in an augmented state, enabling transfer-function and convolutional-kernel analyses. This perspective yields frequency-domain characterizations of memory spectra and clarifies when ESNs emulate structured SSM kernels. Third, we cast teacher forcing as state estimation and propose Kalman/EKF-assisted readout learning, together with EM for hyperparameters (leak, spectral radius, process/measurement noise) and a hybrid subspace procedure for spectral shaping under contraction constraints. 
\end{abstract}

\section{Introduction}
\label{sec:intro}

Reservoir Computing (RC) offers a remarkably frugal route to sequence modeling: by freezing a randomly initialized recurrent operator and training only a linear readout, Echo State Networks (ESNs) achieve competitive modeling capacity with modest computational and statistical burden \cite{jaeger2001echo,maass2002real}. Despite two decades of empirical success and a maturing approximation theory for causal fading–memory maps \cite{BoydChua1985,grigoryeva2018echo,gonon2020riskbounds}, the analytical vocabulary used for ESNs remains partly bespoke—centered on the Echo State Property (ESP), spectral–norm heuristics, and memory–capacity diagnostics—rather than the lingua franca of systems theory. This gap obscures connections to well–developed tools in control, identification, and signal processing, and makes it harder to compare ESNs with the recent wave of state–space sequence models (SSMs) that dominate long–context learning through structured kernels and dissipative dynamics \cite{gu2020hippo,gu2022s4,gu2023mamba}.

This paper puts forward a unifying perspective: \emph{ESNs are naturally expressible as state–space models}. Concretely, the leaky ESN recursion can be written as a nonlinear discrete–time SSM with process and measurement noise,
\begin{equation}
x_{t+1} = (1-\lambda)x_t + \lambda\,\sigma(Wx_t + Uu_t + b) + \omega_t,\quad y_t = Cx_t + \nu_t,
\end{equation}
and, under standard small–signal or operating–point assumptions, admits linear or bilinear SSM surrogates with interpretable poles, gains, and impulse responses. The “leak” parameter $\lambda$ acquires a precise dynamical meaning as a timestep–scaled dissipation factor arising from forward–Euler discretization of a continuous–time ESN ODE/SDE, while readout training corresponds to a classical linear regression (or Bayesian linear estimation) on latent states. Framing ESNs as SSMs imports a powerful arsenal: input–to–state stability (ISS) directly translates and strengthens ESP statements \cite{sontag2008iss,jiang2001iss,manjunath2013}; controllability and observability of lifted or linearized representations offer principled criteria for input scaling and spectral shaping \cite{ljung1999system,vanoverschee1996subspace}; and frequency–domain analysis via transfer functions ties “memory spectra” to pole geometry and convolutional kernels \cite{kailath2000linear}.

Theoretical results on RC already show universality for time–invariant, causal fading–memory filters and precise approximation rates under mixing and Lipschitz assumptions \cite{grigoryeva2018echo,gonon2020riskbounds}. Yet these results are typically phrased without explicit recourse to SSM structure, and ESP is often certified by sufficient conditions such as $\rho(W)<1$ after rescaling or global Lipschitz bounds on $\sigma$. By recasting ESNs within the SSM framework, one can replace ad hoc spectral–radius rules with ISS–based contraction arguments that (i) incorporate input gains $U$ and leaks $\lambda$ in a single small–gain inequality, (ii) quantify fading memory through BIBO/ISS Lyapunov functions rather than scalar heuristics, and (iii) extend to multi–rate and deep (stacked) reservoirs via block–SSM composition with guaranteed stability margins \cite{sontag2008iss,ljung1999system}. In turn, the SSM lens clarifies when and why ESNs emulate modern SSM layers: linearizations and random–feature lifts induce rational transfer functions whose impulse responses are precisely the exponentially decaying kernels exploited by recent S4–type architectures \cite{gu2020hippo,gu2022s4}.

A second payoff of the SSM view is methodological. Teacher forcing can be interpreted as (noisy) state observation; state denoising during training aligns with classical filtering and smoothing \cite{sarkka2013bayesian}. Hyperparameters customarily tuned by grid search—leak $\lambda$, spectral radius/scaling of $W$, input scaling, effective noise levels—admit principled estimation via likelihood maximization or expectation–maximization in SSMs, while subspace identification offers recipes for shaping reservoir spectra subject to ESP/ISS constraints \cite{ljung1999system,vanoverschee1996subspace}. Moreover, Koopman–style lifts supply linear SSMs in expanded coordinates, unifying ESN random features with operator–theoretic embeddings and illuminating the approximation–stability trade–off \cite{mezic2005spectral,williams2015data}. These translations do not merely relabel known ESN practice; they provide a common foundation to compare ESNs, linear RNNs, and structured SSMs, and to reason about generalization through system poles, gains, and dissipativity.

\textbf{Scope and contributions.}  Our contributions are threefold. First, we formalize canonical mappings from ESNs to (i) nonlinear SSMs with exogenous noise, (ii) local linear/bi-linear SSMs via operating–point expansions, and (iii) lifted linear SSMs through random features and Koopman–inspired coordinates. Each mapping comes with explicit conditions under which the induced SSM is well posed and inherits ESP from ISS–type contraction bounds \cite{sontag2008iss,manjunath2013}. Second, we translate core ESN properties into systems language: ESP $\Leftrightarrow$ ISS for the driven recursion; fading memory via Lyapunov and gain–margin certificates; and controllability/observability tests for the linearized/lifted models, connecting reservoir hyperparameters to identifiability and effective memory length \cite{jiang2001iss,ljung1999system}. Third, we develop a frequency–domain account for ESNs: linearizations yield transfer functions $H(z)=C(I-z^{-1}A)^{-1}B$ whose poles quantify memory–accuracy trade–offs and whose impulse responses recover the structured convolutional kernels underlying recent SSM layers \cite{kailath2000linear,gu2022s4}. Throughout, we emphasize how these translations sharpen existing universality theorems and suggest principled design/regularization guidelines without departing from the ESN training paradigm \cite{grigoryeva2018echo,gonon2020riskbounds}.

\textbf{Positioning.} Prior work situates ESNs within RC and nonlinear operator approximation \cite{jaeger2001echo,maass2002real}, while modern sequence models formulate learnable, stable SSMs with long memory via carefully structured kernels \cite{gu2020hippo,gu2022s4,gu2023mamba}. Our thesis is that both lines inhabit the same systems–theoretic landscape. By placing ESNs squarely in SSM territory, we make stability, identifiability, and frequency response first–class analytical objects, enabling apples–to–apples comparisons and cross–fertilization of techniques from classical identification and contemporary deep SSMs \cite{ljung1999system,sarkka2013bayesian}.

\begin{figure}[!ht]
    \centering
    \includegraphics[width=\linewidth]{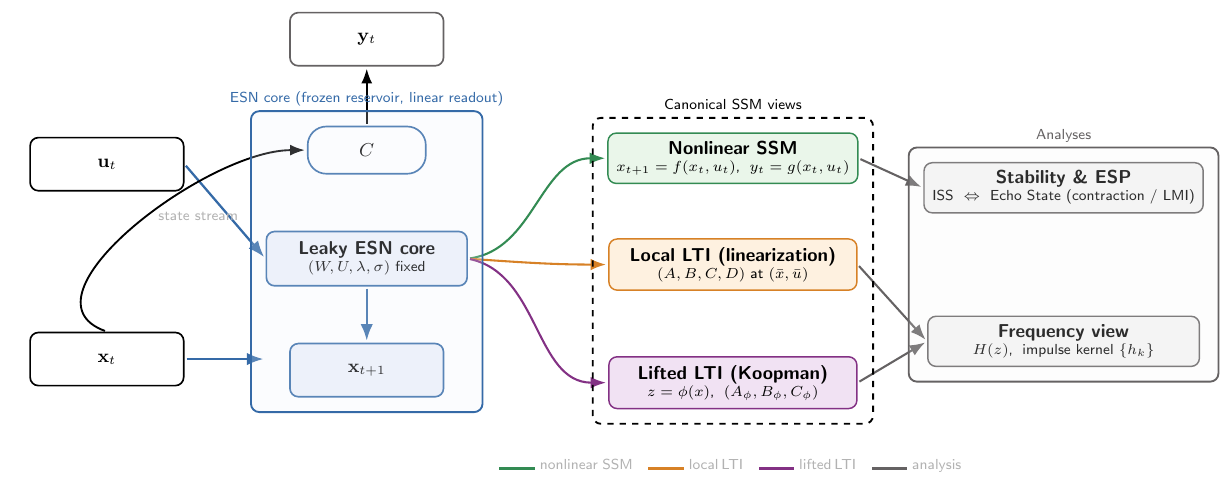}
    \caption{\textbf{ESNs as SSMs: canonical views and analyses.}
Left: a leaky ESN core with fixed reservoir parameters $(W,U,\lambda,\sigma)$ maps $(\mathbf{u}_t,\mathbf{x}_t)\mapsto \mathbf{x}_{t+1}$; a linear readout $C$ yields $\mathbf{y}_t$ (only $C$ is trained).
Middle: the same core is viewed in three canonical SSM forms: (i) a \emph{nonlinear SSM}, $x_{t+1}=f(x_t,u_t),\; y_t=g(x_t,u_t)$; (ii) a \emph{local LTI} model obtained by small–signal linearization at an operating pair, with $(A,B,C,D)$; and (iii) a \emph{lifted/Koopman LTI} model in features $z=\phi(x)$ with $(A_{\phi},B_{\phi},C_{\phi})$.
Right: each view supports a complementary analysis. The nonlinear SSM yields \emph{stability} via input–to–state stability (ISS), which is equivalent to the ESP under contraction or LMI certificates. The LTI views expose the \emph{frequency domain}: transfer function $H(z)=C(I-z^{-1}A)^{-1}B$ and impulse kernel $h_k=CA^{k}B$, whose poles $\lambda_i(A)$ set memory decay $|\lambda_i|$ and oscillation $\arg\lambda_i$.
Together these mappings justify ESN design dials (leak/time constant, spectral scaling, input gain), clarify interpretability (poles/residues $\leftrightarrow$ memory/selectivity), and enable principled identification (KF/EM/subspace) within a unified SSM framework (Secs.~\ref{sec:canonical}–\ref{sec:id}).}

    \label{fig:schematic}
\end{figure}

\textbf{Organization.} Section~\ref{sec:prelim} fixes notation for ESNs and SSMs. Section~\ref{sec:canonical} develops the canonical SSM forms for ESNs (nonlinear, linearized, lifted) and states ISS–based ESP results. Section~\ref{sec:props} analyzes controllability/observability and fading memory in the SSM view. Section~\ref{sec:freq} gives the frequency–domain account and kernel connections. Section~\ref{sec:id} outlines identification perspectives (filtering/EM/subspace) for ESN hyperparameters. Section~\ref{sec:related} situates our perspective in prior work, and Section~\ref{sec:disc} discusses limitations and avenues for structured, multi–rate, and probabilistic reservoirs.

\medskip
\noindent\emph{Remark.} We adopt discrete time as the default analytical setting; continuous–time analogues obtained via standard discretizations (e.g., forward–Euler/Tustin) are noted where relevant.

\section{Preliminaries and Notation}\label{sec:prelim}

Throughout, vectors are column–vectors; $\|\cdot\|$ denotes the Euclidean norm and its induced operator norm for matrices; $\rho(A)$ is the spectral radius of $A$; $I_n$ is the $n{\times}n$ identity. Inputs $u_t\in\mathbb{R}^m$, states $x_t\in\mathbb{R}^n$, and outputs $y_t\in\mathbb{R}^p$. For a sequence $(\xi_t)_{t\in\mathbb{Z}}$, write $\xi_{a:b}=(\xi_a,\ldots,\xi_b)$, and $\ell_\infty(\mathbb{Z}_{\le t};\mathbb{R}^m)$ for bounded semi-infinite input histories. We use “filter” to mean a causal, time-invariant map from input histories to outputs/states.

\subsection{Echo State Networks (leaky update, readout, ESP/fading memory)}\label{sec:prelim-esn}

An Echo State Network (ESN) specifies a driven recurrence with a fixed (random) reservoir and a trained linear readout \cite{jaeger2001echo,maass2002real}. The \emph{leaky} ESN update with affine bias is
\begin{equation}
\label{eq:esn}
x_{t+1}
= (1-\lambda)x_t + \lambda\sigma(Wx_t + Uu_t + b),
\qquad
y_t = Cx_t + d,
\end{equation}
where $W\in\mathbb{R}^{n\times n}$ and $U\in\mathbb{R}^{n\times m}$ are fixed after initialization, $C\in\mathbb{R}^{p\times n}$ (and optionally $d\in\mathbb{R}^p$) are trained, $\sigma:\mathbb{R}^n\to\mathbb{R}^n$ acts componentwise, and $\lambda\in(0,1]$ is the leak. We adopt:

\paragraph{Assumption A1 (activation).}
$\sigma$ is globally Lipschitz with constant $L_\sigma$ (e.g., $\tanh$ has $L_\sigma=1$) and $\sigma(0)=0$.

\paragraph{Remark (continuous–time origin of the leak).}
\eqref{eq:esn} arises from forward–Euler discretization of the ODE
$\tau\dot{x}=-x+\sigma(Wx+Uu+b)$ with step $\Delta t$, identifying $\lambda=\Delta t/\tau$. Thus $\lambda$ controls dissipation/time–scale, a fact we exploit when relating ESNs to SSMs.

\paragraph{Readout learning.}
Given a sequence of states $(x_t)$ and targets $(y_t)$, the readout is typically estimated by ridge regression (or Bayesian linear regression) of $y_t$ on $x_t$ \cite{jaeger2001echo}. Training may use teacher forcing to generate the state trajectory.

\paragraph{Echo State Property.}
Fix bounded inputs $u\in \ell_\infty(\mathbb{Z};\mathbb{R}^m)$. The ESN has the ESP if for every pair of initial states $x_0,x'_0$ the induced trajectories satisfy $\|x_t-x'_t\|\to 0$ as $t\to\infty$; equivalently, for each $t$ there exists a unique state $x_t = \mathcal{X}(u_{-\infty:t})$ that depends causally on the entire past input and is independent of the initial condition \cite{manjunath2013}. A sufficient contraction condition under A1 is
\begin{equation}
\label{eq:esp-cond}
\Big|(1-\lambda)I_n + \lambda WJ_\sigma(\zeta)\Big| \le (1-\lambda) + \lambda |W|L_\sigma < 1,
\end{equation}
uniformly in $\zeta$, which ensures a global input-to-state contraction and thus ESP \cite{manjunath2013}. Classical heuristics (e.g., rescaling to $\rho(W)<1$) are special cases when $L_\sigma\le 1$.

\paragraph{Fading memory.}
Let $d_\alpha(u,v) = \sup_{k\le 0}\alpha^{-k}\|u_k-v_k\|$ for some $\alpha\in(0,1)$ be a weighted metric on histories. A causal filter $F:u_{-\infty:t}\mapsto z_t$ has the \emph{fading memory property} (FMP) if it is continuous from $(\ell_\infty,d_\alpha)$ to $\mathbb{R}^q$, i.e., remote past inputs are exponentially down-weighted. Under \eqref{eq:esp-cond}, the state map $u_{-\infty:t}\mapsto x_t$ and thus the output map $u_{-\infty:t}\mapsto y_t$ have FMP, connecting ESNs to the universality theorems for fading-memory filters \cite{BoydChua1985,grigoryeva2018echo,gonon2020riskbounds}.

\subsection{State Space Models (DT/CT; linear, bilinear, and nonlinear SSMs)}\label{sec:prelim-ssm}

We summarize discrete-time (DT) and continuous-time (CT) state space models in a form that will serve as the receiving framework for ESNs.

\paragraph{Linear time-invariant SSMs (DT/CT).}
The DT LTI model with process/measurement perturbations is
\begin{equation}
\label{eq:lti-dt}
x_{t+1}=Ax_t + Bu_t + w_t,\qquad y_t=Cx_t + Du_t + v_t,
\end{equation}
while the CT counterpart is
\begin{equation}
\label{eq:lti-ct}
\dot{x}(t)=Ax(t)+Bu(t)+w(t),\qquad y(t)=Cx(t)+Du(t)+v(t).
\end{equation}
Here $A\in\mathbb{R}^{n\times n}$, $B\in\mathbb{R}^{n\times m}$, $C\in\mathbb{R}^{p\times n}$, $D\in\mathbb{R}^{p\times m}$. Stability (e.g., $\rho(A)<1$ in DT, $\text{Re}\,\lambda_i(A)<0$ in CT), controllability/observability, and frequency-domain transfer functions $H(z)=C(I-z^{-1}A)^{-1}B + D$ (DT) or $H(s)=C(sI-A)^{-1}B + D$ (CT) are standard \cite{kailath2000linear}. Discretization with step $\Delta t$ yields $A_d=e^{A\Delta t}$, $B_d=\int_0^{\Delta t} e^{A\tau}\,d\tau\, B$, linking \eqref{eq:lti-ct} to \eqref{eq:lti-dt}.

\paragraph{Bilinear SSMs.}
Input–state interactions can be modeled bilinearly:
\begin{equation}
\label{eq:bilinear}
x_{t+1}=Ax_t + Bu_t + \sum_{i=1}^{m} u_{t,i}B_i x_t + w_t,\qquad
y_t=Cx_t + Du_t + v_t,
\end{equation}
or in CT form $\dot{x}=Ax + Bu + \sum_i u_i B_i x + w$. Bilinear systems interpolate between LTI and general nonlinear SSMs while retaining useful analysis tools \cite{kailath2000linear,ljung1999system}.

\paragraph{Nonlinear SSMs and regularity.}
A general SSM is
\begin{equation}
\label{eq:nonlinear}
x_{t+1}=f(x_t,u_t)+w_t,\qquad y_t=g(x_t,u_t)+v_t,
\end{equation}
with $f:\mathbb{R}^n\times\mathbb{R}^m\to\mathbb{R}^n$, $g:\mathbb{R}^n\times\mathbb{R}^m\to\mathbb{R}^p$. Under global Lipschitz or one-sided Lipschitz conditions on $f$ (in $x$) and bounded inputs, input-to-state stability (ISS) yields unique causal filters $u_{-\infty:t}\mapsto x_t$ and fading memory for $y_t$ \cite{sontag2008iss,jiang2001iss}. Linearization of \eqref{eq:nonlinear} about an operating point $(\bar{x},\bar{u})$ gives an LTI surrogate with $A=\partial_x f(\bar{x},\bar{u})$, $B=\partial_u f(\bar{x},\bar{u})$, $C=\partial_x g(\bar{x},\bar{u})$, $D=\partial_u g(\bar{x},\bar{u})$; higher-order terms quantify the validity region.

\paragraph{Noise models and estimation.}
The perturbations $w_t,v_t$ (or $w(t),v(t)$) may be deterministic bounded disturbances or stochastic (e.g., zero-mean with covariances $Q,R$). Classical filtering/smoothing (KF, EKF/UKF in the nonlinear case) and likelihood/EM-based identification operate in \eqref{eq:lti-dt}–\eqref{eq:nonlinear} \cite{ljung1999system,sarkka2013bayesian}. These tools will later instantiate ESN hyperparameter estimation and state denoising in the SSM view.

\paragraph{ESNs as SSMs (preview).}
Equation \eqref{eq:esn} is a special case of \eqref{eq:nonlinear} with $f(x,u)=(1-\lambda)x+\lambda\,\sigma(Wx+Uu+b)$ and $g(x,u)=Cx+d$. Small-signal linearization and random-feature (lifted) coordinates yield LTI surrogates with explicit poles and impulse responses, placing ESNs within the same analytical envelope as \eqref{eq:lti-dt}–\eqref{eq:bilinear}. This identification underpins our use of ISS/ESP, controllability/observability, and frequency-domain tools in subsequent sections.

\subsection{Stability, contractivity, controllability, observability (definitions)}\label{sec:prelim-stability}

We collect the notions used throughout. Unless stated otherwise, all norms are Euclidean and the induced operator norm.

\paragraph{Internal/exponential stability (DT/CT).}
Consider the autonomous DT system $x_{t+1}=f(x_t)$ with equilibrium at the origin.
It is \emph{(globally) exponentially stable} if $\exists\,M\!\ge\!1$, $\alpha\!\in\!(0,1)$ such that for all $x_0$,
\begin{equation}
\|x_t\|\;\le\; M\,\alpha^{\,t}\,\|x_0\|,\qquad \forall t\ge 0.
\end{equation}

For an LTI DT system $x_{t+1}=Ax_t$, exponential stability is equivalent to $\rho(A)<1$.
In CT, $\dot x = f(x)$ is exponentially stable if $\|x(t)\|\le M e^{-\mu t}\|x(0)\|$ for some $M,\mu>0$; for LTI $\dot x = Ax$, this is equivalent to $\max_i \mathrm{Re}\,\lambda_i(A)<0$ \cite{kailath2000linear}.

\paragraph{BIBO stability.}
A causal input–output map is \emph{bounded–input bounded–output} (BIBO) stable if
$\sup_t \|u_t\|<\infty \Rightarrow \sup_t \|y_t\|<\infty$.
For DT LTI $x_{t+1}=Ax_t+Bu_t,\; y_t=Cx_t+Du_t$, BIBO stability holds iff $A$ is DT exponentially stable (as above), equivalently the impulse response is absolutely summable, and the transfer $H(z)=C(I-z^{-1}A)^{-1}B+D$ has no poles on/ outside the unit circle \cite{kailath2000linear}.

\paragraph{Input-to-state stability (ISS).}
For the DT system $x_{t+1}=f(x_t,u_t)$, ISS means there exist comparison functions\footnote{A function $\alpha:\mathbb{R}_{\ge 0}\!\to\!\mathbb{R}_{\ge 0}$ is $\mathcal{K}$ if continuous, strictly increasing, $\alpha(0)=0$; $\mathcal{K}_\infty$ if additionally $\alpha(r)\!\to\!\infty$ as $r\!\to\!\infty$. A function $\beta$ is $\mathcal{KL}$ if $\beta(\cdot,t)\in\mathcal{K}$ for each $t$ and $\beta(r,t)\!\to\!0$ as $t\!\to\!\infty$ for each $r$.}
$\beta\in\mathcal{KL}$ and $\gamma\in\mathcal{K}_\infty$ such that for all initial states and inputs
\begin{equation}
\|x_t\|\;\le\; \beta(\|x_0\|,t) + \gamma\!\Big(\sup_{0\le k<t}\|u_k\|\Big).
\end{equation}

ISS implies the state is a causal, well–posed, fading–memory functional of the input; for LTI it reduces to exponential stability plus a bounded convolution operator. ISS provides the right language for the ESP used in ESNs \cite{sontag2008iss,jiang2001iss,manjunath2013}.

\paragraph{Incremental stability / contraction.}
Let $F_u:\mathbb{R}^n\to\mathbb{R}^n$ be the one-step state map at input $u$, i.e. $x_{t+1}=F_{u_t}(x_t)$.
The system is \emph{contractive} in a norm $\|\cdot\|_\ast$ if there exists $\kappa\in(0,1)$ such that
\begin{equation}
\|F_{u}(x)-F_{u}(x')\|_\ast \;\le\; \kappa\,\|x-x'\|_\ast,\qquad \forall x,x',\;\forall u \text{ in a bounded set}.
\end{equation}

Uniform contraction implies \emph{incremental ISS}: for two trajectories driven by inputs $u,\bar u$,
\begin{equation}
\|x_t-\bar x_t\|_\ast \;\le\; \kappa^t \|x_0-\bar x_0\|_\ast \;+\; \sum_{k=0}^{t-1}\kappa^{t-1-k} L_u \|u_k-\bar u_k\|, 
\end{equation}
for some $L_u$ (Lipschitz in the input). In particular, with identical inputs the difference decays geometrically, which yields uniqueness of the input–driven state (ESP) and exponential fading memory \cite{sontag2008iss,manjunath2013,angeli2002lyapunov}. For ESNs, a sufficient condition is $\|(1-\lambda)I+\lambda WJ_\sigma(\zeta)\|<1$ uniformly (cf. \eqref{eq:esp-cond}).

\paragraph{Controllability (reachability).}
For DT LTI $x_{t+1}=Ax_t+Bu_t$, the pair $(A,B)$ is \emph{controllable} if for any $x_0,x_f$ there exists a finite input sequence steering $x_0$ to $x_f$. Equivalently,
\begin{equation}
\mathrm{rank}\,\mathcal{C} \;=\; n,\qquad 
\mathcal{C}\;:=\;[\,B\; AB\; \cdots\; A^{n-1}B\,].
\end{equation}

If $A$ is DT stable, controllability is also equivalent to the \emph{reachability Gramian}
\begin{equation}
W_c \;=\; \sum_{k=0}^{\infty} A^k B B^\top (A^\top)^k
\end{equation}
being positive definite; $x^\top W_c^{-1}x$ is the minimal input energy to reach $x$ from $0$ \cite{kailath2000linear}.
For nonlinear SSMs $x_{t+1}=f(x_t,u_t)$, \emph{(small-time) local controllability} is defined via reachability of a neighborhood and can be analyzed by the linearization $(A,B)=(\partial_x f,\partial_u f)$ or, more generally, by Lie-algebraic conditions (Hermann–Krener) \cite{hermann1977nonlinear,kailath2000linear}.

\paragraph{Observability.}
For DT LTI $x_{t+1}=Ax_t,\; y_t=Cx_t$, the pair $(A,C)$ is \emph{observable} if the initial state is uniquely determined from a finite output history. Equivalently,
\begin{equation}
\mathrm{rank}\,\mathcal{O} \;=\; n,\qquad 
\mathcal{O}\;:=\;\begin{bmatrix} C \\ CA \\ \vdots \\ CA^{n-1} \end{bmatrix}.
\end{equation}

If $A$ is DT stable, the \emph{observability Gramian}
\begin{equation}
W_o \;=\; \sum_{k=0}^{\infty} (A^\top)^k C^\top C A^k
\end{equation}
is positive definite; $y$-energy $\sum_t \|y_t\|^2$ equals $x_0^\top W_o x_0$ for the zero-input response \cite{kailath2000linear}.
For nonlinear systems, \emph{local weak observability} is defined via the rank of the differential of the \emph{observability map} formed by repeated Lie derivatives of the output along the dynamics (Hermann–Krener rank test) \cite{hermann1977nonlinear}.
In this paper we use these LTI notions primarily for linearized/lifted ESN–as–SSM surrogates, where they provide practical tests and energy interpretations.

\medskip

\begin{table}[!ht]
\centering
\caption{\textbf{Symbols and notation used throughout.}}
\label{tab:notation}
\renewcommand{\arraystretch}{1.12}
\resizebox{\linewidth}{!}{%
\begin{tabular}{@{}lll@{}}
\toprule
\textbf{Symbol} & \textbf{Meaning} & \textbf{Type / Dimensions} \\
\midrule
$t$, $\Delta t$ & discrete time index; step size & $\mathbb{Z}_{\ge 0}$; $\mathbb{R}_{>0}$ \\
$u_t, x_t, y_t$ & input, state, output at time $t$ & $\mathbb{R}^m, \mathbb{R}^n, \mathbb{R}^p$ \\
$w_t, v_t$ & process/measurement perturbations & sequences in $\mathbb{R}^n,\mathbb{R}^p$ \\
$W,U,b$ & ESN reservoir/input weights, bias & $W\!\in\!\mathbb{R}^{n\times n}$, $U\!\in\!\mathbb{R}^{n\times m}$, $b\!\in\!\mathbb{R}^n$ \\
$C,d$ & readout matrix, bias & $C\!\in\!\mathbb{R}^{p\times n}$, $d\!\in\!\mathbb{R}^p$ \\
$\lambda$ & leak (dissipation) parameter & scalar in $(0,1]$ \\
$\sigma$, $J_\sigma$ & activation (componentwise), its Jacobian & $\sigma:\mathbb{R}^n\!\to\!\mathbb{R}^n$; $J_\sigma\in\mathbb{R}^{n\times n}$ \\
$L_\sigma$ & global Lipschitz constant of $\sigma$ & $L_\sigma\in[0,\infty)$ \\
$A,B,C,D$ & SSM/system matrices (LTI) & $A\!\in\!\mathbb{R}^{n\times n}, B\!\in\!\mathbb{R}^{n\times m}, C\!\in\!\mathbb{R}^{p\times n}, D\!\in\!\mathbb{R}^{p\times m}$ \\
$B_i$ & bilinear input–state matrices & $B_i\in\mathbb{R}^{n\times n}$ \\
$\rho(A)$, $\lambda_i(A)$ & spectral radius; eigenvalues of $A$ & $\rho(A)=\max_i|\lambda_i(A)|$ \\
$\|x\|$, $\|M\|$ & Euclidean norm; induced operator norm & $\|M\| = \sup_{\|x\|=1}\|Mx\|$ \\
$\sigma_{\min}(M),\sigma_{\max}(M)$ & extremal singular values & nonnegative scalars \\
$I_n, 0_{a\times b}$ & identity; zero matrix & $n\times n$; $a\times b$ \\
$\langle x,y\rangle$ & inner product & $x^\top y$ \\
$\ell_\infty$ & bounded sequences space & $\{\,\xi: \sup_t \|\xi_t\|\!<\!\infty\,\}$ \\
$\mathcal{K},\mathcal{K}_\infty,\mathcal{KL}$ & comparison function classes & see text (ISS) \\
$H(z), H(s)$ & DT/CT transfer function & matrix–valued rational function \\
$\mathcal{C},\mathcal{O}$ & controllability/observability matrices & $\mathcal{C}=[B\,AB\,\dots\,A^{n-1}B]$; $\mathcal{O}=\begin{bmatrix}C\\CA\\ \vdots\\ CA^{n-1}\end{bmatrix}$ \\
$W_c, W_o$ & reachability/observability Gramians & $W_c=\sum_{k\ge 0}A^kBB^\top (A^\top)^k$, $W_o=\sum_{k\ge 0}(A^\top)^k C^\top C A^k$ \\
$\mathrm{vec}(\cdot)$, $\otimes$ & vectorization; Kronecker product & standard linear–algebra ops \\
$\mathrm{blkdiag}(\cdot)$ & block–diagonal operator & matrix constructor \\
\bottomrule
\end{tabular}}
\end{table}

\smallskip
The symbols in Tab. \ref{tab:notation} are reused verbatim in Sections~\ref{sec:canonical}–\ref{sec:freq}. When a specific norm or inner product different from Euclidean is used (e.g., weighted norms for contraction), this will be indicated explicitly.

\section{Canonical Forms}\label{sec:canonical}

\subsection{ESN as a nonlinear SSM}\label{sec:canonical-nlssm}

We rewrite the leaky ESN as a (possibly stochastic) discrete–time nonlinear state–space model:
\begin{equation}
\label{eq:esn-ssm}
\boxed{~
x_{t+1}=f(x_t,u_t)+\omega_t,\qquad y_t=g(x_t,u_t)+\nu_t,
~}
\end{equation}
with
\begin{equation}
\label{eq:esn-fg}
f(x,u) = (1-\lambda)x + \lambda\sigma(Wx+Uu+b),
\qquad
g(x,u) = Cx + d,
\end{equation}
where $x_t\in\mathbb{R}^n$, $u_t\in\mathbb{R}^m$, $y_t\in\mathbb{R}^p$, $\lambda\in(0,1]$, $\sigma$ acts componentwise, and $\omega_t,\nu_t$ are process/measurement perturbations (deterministic bounded or stochastic).

\paragraph{Regularity and Lipschitz moduli.}
Under Assumption A1 ($\sigma$ globally Lipschitz with constant $L_\sigma$ and $\sigma(0)=0$), $f$ is globally Lipschitz:
\begin{equation}
\|f(x,u)-f(x',u')\|
\;\le\; \underbrace{(1-\lambda)+\lambda\,\|W\|\,L_\sigma}_{=:L_x}\,\|x-x'\|
\;+\; \underbrace{\lambda\,\|U\|\,L_\sigma}_{=:L_u}\,\|u-u'\|.
\end{equation}
Hence, for bounded inputs, \eqref{eq:esn-ssm} is well posed and defines a causal filter.

\paragraph{ISS/ESP in SSM language.}
If $L_x<1$, the one–step map $x\mapsto f(x,u)$ is a contraction uniformly over bounded inputs; standard incremental stability (contraction) then yields input–to–state stability (ISS) and the ESP: the state $x_t$ is the unique causal functional of $u_{-\infty:t}$ independent of $x_0$, with exponential fading memory \cite{sontag2008iss,manjunath2013}. The sufficient condition $L_x<1$ reduces to the usual ESN heuristic $(1-\lambda)+\lambda \|W\|L_\sigma<1$ and implies $\rho\big(\partial_x f(\cdot)\big)\le L_x<1$ (see also Sec.~\ref{sec:prelim-stability}).

\paragraph{Continuous–time provenance.}
Let $\tau\dot{x}=-x+\sigma(Wx+Uu+b)$ and discretize with forward–Euler step $\Delta t$, identifying $\lambda=\Delta t/\tau$. Process noise $\omega_t$ corresponds to integrated CT disturbances; this connects leak $\lambda$ to a dissipation time–scale and will matter when we linearize or lift \eqref{eq:esn-ssm}.

\medskip

\subsection{Small–signal linearization around operating regimes (A, B, C, D; validity domains)}\label{sec:small-signal}

Let $(\bar x,\bar u)$ be an \emph{operating pair} (e.g., an equilibrium satisfying $\bar x=f(\bar x,\bar u)$, or a point on a nominal trajectory $t\mapsto(\bar x_t,\bar u_t)$). Define small deviations $\delta x_t:=x_t-\bar x$ and $\delta u_t:=u_t-\bar u$ (or $\delta x_t:=x_t-\bar x_t$, $\delta u_t:=u_t-\bar u_t$ in the time–varying case). A first–order Taylor expansion of \eqref{eq:esn-ssm} gives the LTI/LTV surrogate
\begin{equation}
\label{eq:lti-ltv}
\delta x_{t+1} = A\delta x_t + B\delta u_t + r_t,
\qquad
\delta y_t = C_y\delta x_t + D\delta u_t + \eta_t,
\end{equation}
with Jacobian blocks evaluated at the operating pair (or along the nominal trajectory)
\begin{equation}
\label{eq:abcd}
\boxed{~
A=(1-\lambda)I + \lambda J_\sigma(\xi)W,\qquad
B=\lambda J_\sigma(\xi)U,\qquad
C_y=\partial_x g = C,\qquad
D=\partial_u g = 0,
~}
\end{equation}
where $\xi:=Wx+Uu+b$ and $J_\sigma(\xi)=\mathrm{diag}(\sigma'(\xi_1),\ldots,\sigma'(\xi_n))$. The remainders $r_t,\eta_t$ collect second–order (and higher) terms.

\paragraph{LTI vs.\ LTV linearization.}
If $(\bar x,\bar u)$ is constant, $(A,B,C_y,D)$ is LTI. If we linearize along a nominal schedule $(\bar x_t,\bar u_t)$, we obtain an LTV model with $A_t=(1-\lambda)I+\lambda J_\sigma(\xi_t)W$ and $B_t=\lambda J_\sigma(\xi_t)U$; the transfer analysis in Sec.~\ref{sec:freq} specializes to frozen–time arguments in that case.

\paragraph{Stability of the linearization.}
$\|A\|\le (1-\lambda)+\lambda \|W\|L_\sigma = L_x$. Thus, under the ESP/ISS sufficient condition $L_x<1$, any frozen–time $A$ is a contraction; in particular, $\rho(A)\le \|A\|<1$, guaranteeing DT exponential stability of the small–signal model (and BIBO stability of its input–output map) \cite{kailath2000linear}.

\paragraph{Validity domains and remainder bounds.}
Assume $\sigma\in C^2$ with uniformly bounded second derivative on a forward–invariant compact set $\mathcal{S}\subset\mathbb{R}^n$: $\|D^2\sigma(z)\|\le L^{(2)}_\sigma$ for all $z\in Wx+Uu+b$ with $(x,u)\in\mathcal{S}\times\mathcal{U}$. Then by Taylor’s theorem,
\begin{equation}
\label{eq:remainder-bound}
|r_t|
\le \frac{\lambda}{2}L^{(2)}_\sigma|W\delta x_t + U\delta u_t|^2,
\qquad
|\eta_t|=0,
\end{equation}
so the linearization error is $O\!\big(\|(\delta x_t,\delta u_t)\|^2\big)$. A convenient validity domain is the tube
\begin{equation}
\mathcal{T}(r)\;:=\;\{(x,u): \|W(x-\bar x)\|+\|U(u-\bar u)\|\le r\},
\end{equation}
on which \eqref{eq:lti-ltv} approximates \eqref{eq:esn-ssm} with one–step error bounded by $\tfrac{\lambda}{2}L^{(2)}_\sigma\, r^2$. In LTV form (linearization along $(\bar x_t,\bar u_t)$), the same bound holds pointwise with $\xi_t=W\bar x_t+U\bar u_t+b$.

\paragraph{Remark (bilinearized surrogates).}
Keeping the first–order dependence of $J_\sigma(\xi)$ on the small perturbations produces a bilinear correction
$\delta x_{t+1}\approx A\,\delta x_t + B\,\delta u_t + \sum_{i}(\delta u_t)_i\,B_i\,\delta x_t$ with $B_i=\lambda\,\mathrm{diag}(\sigma''(\xi)\, (Ue_i))\,W$. This is often unnecessary in practice but clarifies how input–dependent effective dynamics arise \cite{kailath2000linear}.

\medskip

\subsection{Lifted/Koopman SSMs: feature maps to obtain linear models in expanded state}\label{sec:lifted}

The \emph{Koopman operator} acts linearly on observables of a nonlinear dynamical system \cite{mezic2005spectral,rowley2009spectral}. This suggests embedding ESN states into a higher–dimensional \emph{lifted} space where the evolution is approximately linear.

\paragraph{Dictionary of observables and lifted state.}
Choose a feature map (dictionary) $\phi:\mathbb{R}^n\to\mathbb{R}^N$ and define the lifted state $z_t:=\phi(x_t)$. For the ESN map $x_{t+1}=f(x_t,u_t)$, the lifted one–step image is
\begin{equation}
z_{t+1} \;=\; \phi\!\big(f(x_t,u_t)\big).
\end{equation}

If the dictionary is \emph{closed under composition with $f$} (i.e., each component of $\phi\circ f$ lies in the span of $\{1,\phi_1,\ldots,\phi_N\}$ and possibly input features), then there exist matrices $A_\phi,B_\phi,D_\phi$ and vector $c_\phi$ such that the lifted dynamics is \emph{exactly} linear (or bilinear) in $z_t$ and $u_t$:
\begin{equation}
\label{eq:lifted-exact}
z_{t+1} = A_\phi z_t + B_\phi u_t + c_\phi
\quad\text{(or}\ z_{t+1}=A_\phi z_t + B_\phi u_t + \sum_i u_{t,i} B_{\phi,i} z_t + c_\phi\text{)}.
\end{equation}
The output becomes $y_t=C_\phi z_t + d$ with $C_\phi$ selecting the identity component if $\phi$ contains $x$.

\paragraph{Approximate closure: EDMD/Carleman/random features.}
For general analytic $\sigma$, exact closure is unavailable with finite $N$. Three standard approximations yield \eqref{eq:lifted-exact} with bounded residuals:
\begin{enumerate}
\item \emph{Extended DMD (EDMD).} Take a finite dictionary $\phi$ (polynomials, radial basis, Fourier, delays). The Koopman action on $\phi$ is approximated in least squares, yielding $A_\phi,B_\phi,c_\phi$ that minimize the one–step residual $e_t:=\phi(f(x_t,u_t))-(A_\phi\phi(x_t)+B_\phi u_t+c_\phi)$ \cite{williams2015data,rowley2009spectral}. In theory sections, we assume access to such a representation with a known uniform bound $\|e_t\|\le \varepsilon$ on a compact forward–invariant set.\footnote{We analyze the consequences of such a bound; empirical identification of $A_\phi,B_\phi,c_\phi$ is orthogonal to our theoretical claims.}
\item \emph{Carleman lifting.} Expand the analytic nonlinearity via Taylor series and collect monomials of $x$ (and $u$) up to degree $d$. The augmented monomial vector $\psi_d(x,u)$ evolves under a \emph{block upper–triangular} linear operator; truncation at degree $d$ yields a linear model with remainder controlled by the tail of the series on a compact domain \cite{kowalski1991nonlinear}. For the ESN, expanding $\sigma(\xi)=\sum_{k=0}^\infty a_k \xi^k$ around an operating point makes $\xi=Wx+Uu+b$ polynomial in $(x,u)$, hence Carleman applies.
\item \emph{Random/kernal features.} For shift–invariant kernels, random Fourier features approximate nonlinear maps by finite–dimensional features $\phi(x)=\sqrt{2/N}[\cos(\omega_i^\top x+b_i)]_{i=1}^N$. One can then regress a linear $A_\phi,B_\phi$ so that $\phi\circ f\approx A_\phi\phi+B_\phi u+c_\phi$ on a compact set, with uniform error controlled by kernel approximation plus regression error (Stone–Weierstrass–type arguments apply to rich dictionaries).
\end{enumerate}

\paragraph{A generic lifted surrogate and its guarantees.}
Assume a compact forward–invariant set $\mathcal{S}\times\mathcal{U}$ and a dictionary $\phi$ containing a constant and the identity, such that
\begin{equation}
\label{eq:lift-approx}
\sup_{(x,u)\in\mathcal{S}\times\mathcal{U}}
\Big|\phi\big(f(x,u)\big) - \big(A_\phi\phi(x) + B_\phi u + c_\phi\big)\Big|
\le \varepsilon.
\end{equation}
Then the lifted LTI SSM
\begin{equation}
\label{eq:lifted-ssm}
z_{t+1}=A_\phi z_t + B_\phi u_t + c_\phi + \tilde\omega_t,\qquad
y_t=C_\phi z_t + d + \tilde\nu_t,
\end{equation}
with $z_t=\phi(x_t)$, approximates the ESN on $\mathcal{S}\times\mathcal{U}$ with one–step model error $\|\tilde\omega_t\|\le \varepsilon$. If $\rho(A_\phi)<1$, the lifted surrogate is DT exponentially stable and BIBO stable; standard small–gain arguments imply an \emph{a priori} bound on the $k$-step prediction error that contracts at rate $\rho(A_\phi)$ and accumulates linearly in $\varepsilon$ (geometric series).

\paragraph{Connecting to frequency–domain kernels.}
When $\phi$ includes identity and low–order monomials of preactivations $\xi=Wx+Uu+b$, the block structure of $A_\phi$ is (nearly) upper triangular with a stable diagonal block equal to the small–signal $A$ from \eqref{eq:abcd}. The induced transfer function
$H(z)=C_\phi\,(I - z^{-1}A_\phi)^{-1} B_\phi + D_\phi$
has exponentially decaying impulse responses whose time constants are controlled by the eigenvalues of $A_\phi$. This makes precise how ESNs, when viewed through linearizations or lifts, implement the structured convolutional kernels exploited by modern SSM layers \cite{gu2020hippo,gu2022s4}.

\paragraph{Remarks.}
(i) If $\phi$ augments the identity with time–delay coordinates (Hankel features), then \eqref{eq:lifted-ssm} recovers classical finite–impulse–response approximations to fading–memory filters within a linear state–space \cite{kailath2000linear}.
(ii) Stability of the base ESN (ISS/ESP) aids the lift: if $x_t$ stays in a compact $\mathcal{S}$ for bounded inputs, then $\phi$ and the Taylor/EDMD residuals are uniformly controlled on $\mathcal{S}$.

\subsection{Random-feature SSMs}\label{sec:randfeat-ssm}

Consider the leaky ESN map
\begin{equation}
x_{t+1}=(1-\lambda)x_t+\lambda\,\sigma(\psi_t)+\omega_t,\qquad 
\psi_t:=Wx_t+Uu_t+b,\qquad y_t=Cx_t+\nu_t.
\end{equation}

For each coordinate, approximate the static nonlinearity $\sigma:\mathbb{R}\!\to\!\mathbb{R}$ on a compact set $\Psi\subset\mathbb{R}$ by a fixed (data-independent) feature dictionary $\phi:\mathbb{R}\!\to\!\mathbb{R}^{N}$ and a linear combiner:
\begin{equation}
\label{eq:sigma-rf-scalar}
\sigma(\xi)\approx v^\top \phi(\xi),\qquad \sup_{\xi\in\Psi}\big|\sigma(\xi)-v^\top\phi(\xi)\big|\le \varepsilon,
\end{equation}
with $v\in\mathbb{R}^{N}$. Choices include random Fourier features for shift-invariant kernels, random ridge features, or bounded orthonormal systems; approximation rates of order $O_\mathbb{P}(1/\sqrt{N})$ are classical under RKHS/Barron-type regularity \cite{rahimi2008randomfeatures,rahimi2009weighted,barron1993universal,bach2017breaking}. Stacking these approximations componentwise yields a matrix $V\in\mathbb{R}^{n\times N}$ and a vector feature map $\Phi:\mathbb{R}^n\!\to\!\mathbb{R}^{N}$ such that
\begin{equation}
\label{eq:vector-rf}
\sigma(\psi)\approx V,\Phi(\psi),\qquad
\sup_{\psi\in\Psi^n}|\sigma(\psi)-V\Phi(\psi)|\le \varepsilon.
\end{equation}

\paragraph{Lur’e form and “augmented LTI core.”}
Substituting \eqref{eq:vector-rf} gives
\begin{equation}
\label{eq:lure}
x_{t+1}=A_0 x_t + B_u u_t + B_z z_t + \omega_t,\qquad
z_t=\Phi(G_x x_t + G_u u_t + g_0),\qquad y_t=Cx_t+\nu_t,
\end{equation}
with $A_0=(1-\lambda)I$, $B_u=0$ (or a direct term if desired), $B_z=\lambda V$, $G_x=W$, $G_u=U$, $g_0=b$. The pair $(A_0,B_z)$ defines a \emph{linear time-invariant core} fed back through the \emph{static, memoryless} nonlinearity $z=\Phi(\cdot)$; this is the canonical Lur’e interconnection. If $\Phi$ is globally Lipschitz on $\Psi^n$ with constant $L_\Phi$, the closed loop inherits ISS/ESP from small-gain (circle-criterion/IQC) conditions.

\paragraph{A sufficient small-gain condition (robust to RF error).}
Let $L_\Phi:=\sup_{\xi\ne\xi'}\|\Phi(\xi)-\Phi(\xi')\|/\|\xi-\xi'\|$ on $\Psi^n$. Then for the idealized model (no RF error),
\begin{equation}
\|x_{t+1}-\bar x_{t+1}\|\le \|A_0\|\|x_t-\bar x_t\|+\|B_z\|\,\|\Phi(G_x x_t+G_u u_t+g_0)-\Phi(G_x \bar x_t+G_u u_t+g_0)\|
\end{equation}
\begin{equation}
\le \Big(\|A_0\|+\|B_z\|\,L_\Phi\,\|G_x\|\Big)\|x_t-\bar x_t\|.
\end{equation}

Hence a uniform contraction (and thus ESP/ISS) holds if
\begin{equation}
\label{eq:rf-smallgain}
\boxed{\qquad |A_0|+|B_z|L_\Phi|G_x| < 1 \quad\Longleftrightarrow\quad (1-\lambda)+\lambda|V|L_\Phi|W|<1. \qquad}
\end{equation}
When the RF approximation incurs uniform error $\varepsilon$ as in \eqref{eq:vector-rf}, the one-step model mismatch acts as an additive disturbance $\tilde\omega_t$ with $\|\tilde\omega_t\|\le \lambda\,\varepsilon$. Under \eqref{eq:rf-smallgain}, the induced state error remains bounded by a geometric series, yielding an ISS bound with ultimate radius $O(\lambda\varepsilon/(1-\kappa))$, where $\kappa:=(1-\lambda)+\lambda\|V\|L_\Phi\|W\|$ \cite{sontag2008iss,angeli2002lyapunov}.

\paragraph{LMI (quadratic) certificate via circle criterion.}
Suppose $\Phi$ is sector-bounded/Lipschitz: $\|\Phi(\xi)-\Phi(\xi')\|\le L_\Phi\|\xi-\xi'\|$. A quadratic Lyapunov certificate for absolute stability of \eqref{eq:lure} is given by $P\succ0$ satisfying the LMI (one formulation)
\begin{equation}
\label{eq:lmi}
\begin{bmatrix}
A_0^\top P A_0 - P & A_0^\top P B_z \\
B_z^\top P A_0 & B_z^\top P B_z - \tfrac{1}{L_\Phi^{2}},I
\end{bmatrix}\prec 0,
\end{equation}
which implies a contraction in the weighted norm $\|x\|_P:=\sqrt{x^\top P x}$ for the feedback interconnection (discrete-time circle criterion/IQC; see, e.g., \cite{khalil2002nonlinear,megretski1997iqc}). This yields an ESP/ISS guarantee that can be tuned via $\lambda$, $\|V\|$, and $\|W\|$.

\paragraph{Remarks.}
(i) The random-feature view keeps the dynamics linear in the \emph{unknown} parameters while isolating nonlinearity in a fixed, memoryless block; it is thus complementary to Koopman lifts (Sec.~\ref{sec:lifted}).
(ii) If $\Phi$ contains the identity and low-order monomials of preactivations, one recovers the small-signal matrices $A=(1-\lambda)I+\lambda J_\sigma(\xi)W$, $B=\lambda J_\sigma(\xi)U$ as first-order terms.
(iii) For design, \eqref{eq:rf-smallgain} suggests trading off $\lambda$ and $\|V\|\,\|W\|$ to target a memory timescale $1/(1-\kappa)$ while preserving ESP.

\medskip

\subsection{Continuous-time ESN as ODE/SDE and its discretization}\label{sec:ct-disc}

A continuous-time (CT) ESN with time constant $\tau>0$ is the first-order lag driven by a static nonlinearity:
\begin{equation}
\label{eq:ct-esn-ode}
\tau\dot x(t)= -x(t)+\sigma\big(Wx(t)+Uu(t)+b\big)+w_c(t),\qquad y(t)=C x(t)+v_c(t),
\end{equation}
where $w_c,v_c$ are (deterministic bounded) disturbances or, in an SDE model, white-noise terms:\footnote{SDE form: $dx = \tfrac{1}{\tau}\big(-x+\sigma(\cdot)\big)\,dt + \Sigma\,dB_t$, with $B_t$ a Wiener process and diffusion $\Sigma$.}
\begin{equation}
dx(t)=\frac{1}{\tau}\!\Big(-x(t)+\sigma(\psi(t))\Big)\,dt+\Sigma\,dB_t,\qquad \psi(t)=Wx(t)+Uu(t)+b.
\end{equation}

\paragraph{Explicit-Euler (forward-Euler) discretization.}
Sampling with period $\Delta t$ and zero-order hold on $u$ yields the Euler–Maruyama scheme \cite{higham2001sde,kloeden1992sde}:
\begin{align}
x_{t+1}&=x_t+\frac{\Delta t}{\tau}\Big(-x_t+\sigma(Wx_t+Uu_t+b)\Big)+\omega_t,\\[2pt]
&=(1-\lambda)x_t+\lambda\sigma(Wx_t+Uu_t+b)+\omega_t,\qquad
\boxed{\ \lambda:=\frac{\Delta t}{\tau}\ },
\label{eq:euler-leak}
\end{align}
with $\omega_t\sim\mathcal{N}(0,Q_d)$ in the SDE case, where $Q_d=\Delta t\,\Sigma\Sigma^\top$ (to first order). Thus the discrete-time leak $\lambda$ is the Euler step-normalized inverse time constant. For $\Delta t\ll \tau$, this coincides with the “leaky ESN” update used throughout.

\paragraph{Exact discretization of the linearized CT model.}
Linearize \eqref{eq:ct-esn-ode} around an operating pair $(\bar x,\bar u)$. Writing $f_c(x,u):=\tfrac{1}{\tau}\big(-x+\sigma(Wx+Uu+b)\big)$, the linearization has matrices
\begin{equation}
A_c:=\partial_x f_c(\bar x,\bar u)=\tfrac{1}{\tau}\Big(-I+J_\sigma(\xi)\,W\Big),\qquad 
B_c:=\partial_u f_c(\bar x,\bar u)=\tfrac{1}{\tau}\,J_\sigma(\xi)\,U,\quad \xi=W\bar x+U\bar u+b.
\end{equation}

Under zero-order hold, the exact DT equivalent is \cite{kailath2000linear,sarkka2013bayesian}
\begin{equation}
\label{eq:exact-disc}
A_d=e^{A_c\Delta t},\qquad
B_d=\Big(\int_{0}^{\Delta t} e^{A_c \tau}d\tau\Big)B_c,\qquad
Q_d=\int_{0}^{\Delta t} e^{A_c \tau} Q_c\ e^{A_c^\top \tau} d\tau,
\end{equation}
with $Q_c=\Sigma\Sigma^\top$ the CT diffusion. For small $\Delta t$, $A_d=I+\Delta t A_c+O(\Delta t^2)$ and $B_d=\Delta t B_c+O(\Delta t^2)$, recovering Euler.

\paragraph{Bilinear/Tustin (trapezoidal) discretization.}
An alternative numerically stable mapping (the bilinear or Tustin transform) applied to the scalar lag $\tau \dot x=-x + s(t)$ gives the DT update
\begin{equation}
x_{t+1} = a_{\text{Tustin}}\, x_t + (1-a_{\text{Tustin}})\, s_t,\qquad 
a_{\text{Tustin}}=\frac{1-\alpha}{1+\alpha},\quad \alpha:=\frac{\Delta t}{2\tau}.
\end{equation}

In the ESN, $s_t=\sigma(Wx_t+Uu_t+b)$. Hence the effective “leak” is
\begin{equation}
\label{eq:tustin-leak}
\lambda_{\text{Tustin}} := 1-a_{\text{Tustin}} = \frac{2\alpha}{1+\alpha}
= \frac{\Delta t/\tau}{1+\tfrac{\Delta t}{2\tau}}
= \frac{\Delta t}{\tau+\tfrac{\Delta t}{2}}.
\end{equation}
For $\Delta t\ll \tau$, $\lambda_{\text{Tustin}}\approx \Delta t/\tau=\lambda$; for larger $\Delta t$, Tustin yields a slightly smaller effective leak (more damping) and improved numerical robustness \cite{franklin2015feedback}.

\paragraph{ISS/ESP in CT and preservation under discretization.}
If $\sigma$ is globally Lipschitz with constant $L_\sigma$, then $f_c$ is one-sided Lipschitz with constant $\mu_c=\tfrac{1}{\tau}(-1+L_\sigma\|W\|)$. Whenever $-1+L_\sigma\|W\|<0$ (i.e., $L_\sigma\|W\|<1$), the CT system is incrementally exponentially stable (hence ISS) for bounded inputs; explicit Euler with sufficiently small $\Delta t$ preserves incremental stability (and thus ESP) because the DT contraction factor satisfies
\begin{equation}
\|A_d\|\le (1-\lambda)+\lambda\,L_\sigma\|W\|\;=\;1-\tfrac{\Delta t}{\tau}\big(1-L_\sigma\|W\|\big)\;<\;1,
\end{equation}
matching the discrete small-gain condition in Secs.~\ref{sec:prelim-esn}–\ref{sec:canonical-nlssm}.

\paragraph{Noise mapping (SDE $\to$ DT).}
In the linearized SDE, the DT process covariance is the Lyapunov integral in \eqref{eq:exact-disc}. To first order, $Q_d\approx Q_c\,\Delta t$. This scaling underlies the common practice of tying the ESN process noise magnitude to the sampling period in probabilistic formulations \cite{sarkka2013bayesian}.

\paragraph{Takeaways.}
(i) The leaky ESN is the explicit-Euler discretization of a simple CT lag; the “leak” is $\lambda=\Delta t/\tau$ to first order.
(ii) Exact (linearized) discretization and Tustin yield alternative $\lambda$–$\Delta t$ relations \eqref{eq:exact-disc}–\eqref{eq:tustin-leak} with better numerical damping at larger steps.
(iii) ISS/ESP conditions match between CT and DT descriptions when expressed in terms of dissipation $1/\tau$, Lipschitz slope $L_\sigma\|W\|$, and step $\Delta t$.

\section{System-Theoretic Properties of ESN-as-SSM}\label{sec:props}

We work with the nonlinear SSM formulation of the leaky ESN from \eqref{eq:esn-ssm}–\eqref{eq:esn-fg}, under Assumption~A1 ($\sigma$ globally Lipschitz with constant $L_\sigma$, $\sigma(0)=0$). Norms are Euclidean unless stated otherwise.

\subsection{Echo State Property as input-to-state stability}\label{sec:props-esp-iss}

\paragraph{Setup.}
Consider $x_{t+1}=f(x_t,u_t)$ with $f(x,u)=(1-\lambda)x+\lambda\sigma(Wx+Uu+b)$. Let $\mathcal{U}_M:=\{u: \sup_t\|u_t\|\le M\}$ be a bounded input class. Recall: ISS means $\exists\;\beta\!\in\!\mathcal{KL},\,\gamma\!\in\!\mathcal{K}_\infty$ s.t.
\begin{equation}\label{eq:iss}
|x_t|\le\beta(|x_0|,t)+\gamma\Big(\sup_{0\le k<t}|u_k|\Big)\quad \forall t,
\end{equation}
and \emph{incremental ISS} (iISS) means for any two solutions under inputs $u,\bar u$,
\begin{equation}\label{eq:iiss}
|x_t-\bar x_t|\le \kappa^t|x_0-\bar x_0|+\sum_{k=0}^{t-1}\kappa^{t-1-k}L_u|u_k-\bar u_k|,
\end{equation}
for some $\kappa\!\in\!(0,1)$, $L_u\!\ge 0$. The ESP asserts: for every $u\in\mathcal{U}_M$ there exists a unique entire state trajectory $(x_t)$ that is a causal functional of $u_{-\infty:t}$ and independent of $x_0$ \cite{manjunath2013}.

\begin{proposition}[ESP $\Leftrightarrow$ (incremental) ISS for ESN]\label{prop:esp-iss}
Suppose $f$ is globally Lipschitz in $x$ with constant $L_x<1$ and Lipschitz in $u$ with constant $L_u$ (Sec.~\ref{sec:canonical-nlssm}). Then:
(i) the map $x\mapsto f(x,u)$ is a global contraction uniformly over bounded $u$; hence incremental ISS \eqref{eq:iiss} holds with $\kappa=L_x$;
(ii) for each $u\in\mathcal{U}_M$ there exists a unique input-driven solution $x_t=\mathcal{X}(u_{-\infty:t})$ satisfying ESP;
(iii) the state and output maps have exponential fading memory and \eqref{eq:iss} holds (ISS).
Conversely, if ESP holds uniformly over $\mathcal{U}_M$ and $f$ is continuous and globally Lipschitz in $u$, then there exists an equivalent norm $\|\cdot\|_\ast$ under which $f(\cdot,u)$ is a contraction with a common constant $\kappa<1$ on compact forward-invariant sets, implying iISS.
\end{proposition}

\emph{Sketch.} The forward implication follows from standard contraction/iISS arguments (Banach fixed-point theorem on the state-update operator and telescoping differences) and yields ESP plus fading memory \cite{sontag2008iss,angeli2002lyapunov,manjunath2013}. The reverse direction uses equivalence of norms on finite-dimensional spaces and uniform attractivity of the input-driven solution to construct an ISS-Lyapunov function and a contracting metric \cite{sontag2008iss}. $\square$

\paragraph{Constants for the ESN.}
With Assumption~A1,
\begin{equation}
\|f(x,u)-f(x',u')\|\;\le\; \underbrace{(1-\lambda)+\lambda\,\|W\|\,L_\sigma}_{=:L_x}\,\|x-x'\|\;+\;\underbrace{\lambda\,\|U\|\,L_\sigma}_{=:L_u}\,\|u-u'\|.
\end{equation}

Thus $L_x<1$ is a sufficient global condition for ESP/ISS; see also the weighted-norm generalization in Sec.~\ref{sec:props-contr}.

\subsection{Sufficient conditions via Lipschitz constants and spectral radius}\label{sec:props-contr}

We collect three practically checkable sufficient conditions ensuring contraction (hence ESP/ISS).

\paragraph{(C1) Global Lipschitz contraction.}
As above,
\begin{equation}\label{eq:C1}
(1-\lambda)+\lambda|W|L_\sigma<1 \quad\Longrightarrow\quad \text{ESP and ISS with rate }\kappa=L_x.
\end{equation}
This is norm-dependent but conservative; it transparently links leak $\lambda$, reservoir gain $\|W\|$, and nonlinearity slope $L_\sigma$.

\paragraph{(C2) Weighted (quadratic) contraction via an LMI.}
Let $J_\sigma(\xi)$ be the diagonal Jacobian with $\|J_\sigma(\xi)\|\le L_\sigma$ for all $\xi$. Define the Jacobian of the one-step map in $x$:
\begin{equation}
\mathcal{J}(x,u)\;=\;(1-\lambda)I+\lambda\,J_\sigma(Wx+Uu+b)\,W.
\end{equation}

If there exists $P\succ 0$ and $\kappa\in(0,1)$ such that the uniform matrix inequality
\begin{equation}\label{eq:C2}
\mathcal{J}(x,u)^\top P\mathcal{J}(x,u)\preceq \kappa^2P\qquad \text{for all } (x,u)\text{ in a forward-invariant set},
\end{equation}
holds, then $\|x\|_P:=\sqrt{x^\top P x}$ is a contracting metric and ESP/ISS follow with rate $\kappa$. A sufficient (conservative) LMI independent of $(x,u)$ is
\begin{equation}\label{eq:C2LMI}
\Big((1-\lambda)I+\lambda\Delta W\Big)^\top P \Big((1-\lambda)I+\lambda\Delta W\Big)\preceq \kappa^2 P
\quad\text{for all diagonal }\Delta,~|\Delta|\le L_\sigma,
\end{equation}
obtainable by S-procedure/IQC bounds for slope-restricted $\sigma$ (discrete-time circle criterion) \cite{khalil2002nonlinear,megretski1997iqc}.

\paragraph{(C3) Local small-signal (spectral-radius) condition.}
At an operating pair $(\bar x,\bar u)$ (or along a nominal trajectory), the small-signal model has
\begin{equation}
A=(1-\lambda)I+\lambda\,J_\sigma(\xi)\,W,\qquad \xi=W\bar x+U\bar u+b.
\end{equation}

If $\rho(A)<1$ (equivalently, there exists $P\succ0$ with $A^\top P A\prec P$), the linearized dynamics is DT exponentially stable; this certifies local (tube) contraction of the nonlinear map and local ESP with an $O(\|(\delta x,\delta u)\|^2)$ remainder (Sec.~\ref{sec:small-signal}; \cite{kailath2000linear}). A conservative but convenient bound is $\rho(A)\le \|A\|\le (1-\lambda)+\lambda\,L_\sigma\|W\|$.

\paragraph{Input gains and iISS.}
Under any of (C1)–(C3), the input Lipschitz constant $L_u=\lambda\,\|U\|\,L_\sigma$ yields the iISS bound \eqref{eq:iiss}. This makes explicit the memory-versus-gain trade-off: as $\kappa\uparrow 1$, the effective horizon grows (Sec.~\ref{sec:props-fmp}), but sensitivity to input perturbations increases proportionally to $1/(1-\kappa)$.

\subsection{Controllability/observability of the lifted (linear) system}\label{sec:props-co}

We analyze the lifted surrogate from Sec.~\ref{sec:lifted}:
\begin{equation}\label{eq:liftLTI}
z_{t+1}=A_\phi z_t + B_\phi u_t + c_\phi,\qquad y_t=C_\phi z_t + d,
\end{equation}
(neglecting bounded approximation error for exposition). These notions govern identifiability and the effectiveness of readouts on lifted states.

\paragraph{LTI rank tests.}
For an LTI surrogate with stable $A_\phi\in\mathbb{R}^{N\times N}$:
\begin{equation}
\text{Controllability: } \operatorname{rank}\,\mathcal{C}_\phi = N,\quad 
\mathcal{C}_\phi := [\,B_\phi\; A_\phi B_\phi\; \cdots\; A_\phi^{N-1}B_\phi\,],
\end{equation}
\begin{equation}
\text{Observability: } \operatorname{rank}\,\mathcal{O}_\phi = N,\quad 
\mathcal{O}_\phi := \begin{bmatrix} C_\phi \\ C_\phi A_\phi \\ \vdots \\ C_\phi A_\phi^{N-1}\end{bmatrix}.
\end{equation}

Equivalently, Gramians
\begin{equation}
W_c=\sum_{k=0}^{\infty}A_\phi^k B_\phi B_\phi^\top (A_\phi^\top)^k\succ0,\qquad
W_o=\sum_{k=0}^{\infty}(A_\phi^\top)^k C_\phi^\top C_\phi A_\phi^k\succ0,
\end{equation}
are positive definite \cite{kailath2000linear}.

\paragraph{Practical criteria for large $N$.}
(i) \emph{Truncated Gramians.} For $\rho(A_\phi)<1$, $\|A_\phi^k\|\le c\,\rho(A_\phi)^k$; thus finite truncations
$W_c^{(T)}:=\sum_{k=0}^{T}A_\phi^k B_\phi B_\phi^\top (A_\phi^\top)^k$
converge geometrically to $W_c$. In practice one tests $\lambda_{\min}(W_c^{(T)})> \epsilon$ for moderate $T$ (e.g., $T\!\approx\!\lceil 5/(1-\rho(A_\phi))\rceil$).
(ii) \emph{Persistent excitation.} Identification and controllability require inputs that are \emph{persistently exciting of order $N$}: there exists $\alpha>0$ such that the block Toeplitz input covariance over a window $T\ge N$ is $\succeq \alpha I$ \cite{ljung1999system,vanoverschee1996subspace}. This ensures $W_c^{(T)}\succ 0$ empirically.
(iii) \emph{Dictionary design.} If $\phi$ includes the identity on $x$ and $C_\phi$ selects those coordinates, then observability of the base block is immediate; otherwise one must ensure that the chosen readout $C_\phi$ “sees” all dynamically active directions (e.g., include identity and low-order monomials of $x$ or preactivations $\xi$).
(iv) \emph{Numerical conditioning.} Even when rank tests pass, ill-conditioned $W_o,W_c$ imply weak identifiability and large variance in readouts/regression. Regularization (ridge, priors) or structural constraints on $A_\phi$ (normal/banded/diagonal-plus-low-rank) mitigate this.

\paragraph{Local analysis via small-signal blocks.}
In the small-signal LTI model $(A,B,C)$ of Sec.~\ref{sec:small-signal}, controllability/observability of $(A,B)$, $(A,C)$ give local criteria for how input scaling ($U$) and readout placement ($C$) affect excitation and identifiability. In particular, if $A$ is close to normal with eigenvalues inside the unit disk, $W_o$ concentrates along eigen-directions with $|\lambda_i(A)|$ near 1; these directions carry long memory and are easiest to observe \cite{kailath2000linear}.

\subsection{Fading memory and effective horizon; links to eigenvalues and mixing}\label{sec:props-fmp}

\paragraph{Fading memory via contraction.}
Under iISS with contraction factor $\kappa\in(0,1)$ and input Lipschitz $L_u$, a perturbation $\delta u_t$ at time $t\!-\!h$ affects $x_t$ by at most $\kappa^{h-1}L_u\|\delta u_{t-h}\|$. Fix $\varepsilon>0$ and bound $\|\delta u\|\le M$. The \emph{effective memory horizon} $H_\varepsilon$ (the smallest $h$ so that contributions older than $h$ are below $\varepsilon$) satisfies
\begin{equation}\label{eq:Heps}
H_\varepsilon \le \left\lceil \frac{\log\big(\tfrac{L_uM}{\varepsilon}\big)}{-\log \kappa} \right\rceil
= \left\lceil \frac{\log(L_u M/\varepsilon)}{1-\kappa+O((1-\kappa)^2)} \right\rceil,
\end{equation}
so $H_\varepsilon$ scales like $1/(1-\kappa)$ near the edge of stability. This provides a quantitative design rule linking leak/spectral scaling ($\kappa$) to memory length \cite{sontag2008iss,BoydChua1985}.

\paragraph{Frequency-domain view (linearized/lifted).}
For a stable LTI/LTI-lifted surrogate, $y_t=\sum_{k\ge 0}h_k u_{t-k}$ with impulse blocks $h_k=CA^{k}B$. Then
$\|h_k\|\le \|C\|\,\|A^k\|\,\|B\|\le c\,\|C\|\,\|B\|\,\rho(A)^k,$
so memory decay is governed by $\rho(A)$. If $A$ is diagonalizable $A=V\Lambda V^{-1}$, then
$h_k = C V \Lambda^k V^{-1} B = \sum_i \lambda_i^k\,C v_i w_i^\top B$
revealing oscillatory memory (complex $\lambda_i=re^{j\omega}$) with envelope $r^k$. Directions with $|\lambda_i|\approx 1$ dominate long-horizon responses and contribute most to observability/controllability Gramians \cite{kailath2000linear}. This explains the kernel shapes used in modern SSMs (exponentially decaying/oscillatory) and their ESN counterparts (Sec.~\ref{sec:freq}; \cite{gu2022s4}).

\paragraph{Stochastic inputs/noise and mixing.}
With bounded Lipschitz $f$ and additive disturbances/noise ($\omega_t$) or random inputs $u_t$, the state process $(x_t)$ is a time-homogeneous Markov chain on $\mathbb{R}^n$ with a global contraction in a Wasserstein (or weighted Euclidean) metric when $\kappa<1$. Standard results imply geometric ergodicity and geometric decay of mixing coefficients at rate $\kappa$ (up to constants), i.e., $\beta(h)\le C\,\kappa^h$ \cite[Ch.~15]{meyn2009markov}. Thus the deterministic fading-memory rate and the stochastic mixing rate coincide, tying effective horizon to statistical dependence decay.

\paragraph{Memory capacity heuristics.}
For scalar LTI surrogates $x_{t+1}=a x_t+ b u_t$, the contribution of $u_{t-h}$ to $y_t$ scales like $a^h$, yielding the classical “memory capacity” growth as $a\uparrow 1$ (edge-of-stability) but with increased sensitivity to noise/perturbations—precisely the $1/(1-\kappa)$ trade-off that appears in \eqref{eq:Heps} \cite{BoydChua1985}. (For information-theoretic capacity measures in RC, see \cite{Dambre2012}.)

\medskip

The properties above—ESP as ISS, contraction-based sufficient conditions, CO tests on lifted surrogates, and quantitative memory horizons—provide a unified, systems-theoretic foundation for ESNs. They justify principled design dials (leak $\lambda$, spectral scaling $\|W\|$, input gain $\|U\|$) and clarify trade-offs between long memory, stability margins, and identifiability.

\section{Frequency–Domain View and Convolutional Kernels}\label{sec:freq}

We analyze the (linearized or lifted) ESN–as–SSM in the frequency domain and relate its impulse kernels to modern SSM convolutions. Throughout this section, $A\in\mathbb{R}^{N\times N}$, $B\in\mathbb{R}^{N\times m}$, $C\in\mathbb{R}^{p\times N}$ denote a stable LTI surrogate obtained either by small–signal linearization (Sec.~~\ref{sec:small-signal}) or by a lifted representation (Sec.~~\ref{sec:lifted}); $D$ is an optional feedthrough (zero for the standard ESN readout).

\subsection{Transfer function (linearized/lifted case)}\label{sec:freq-transfer}

For the discrete–time LTI model
\begin{equation}\label{eq:lti}
x_{t+1}=Ax_t+Bu_t,\qquad y_t=Cx_t+Du_t,
\end{equation}
the (matrix) \emph{transfer function} is the $z$–transform of the impulse response \cite{kailath2000linear}:
\begin{equation}\label{eq:transfer}
\boxed{ H(z) = C(I-z^{-1}A)^{-1}B + D = C\Big(\sum_{k=0}^\infty z^{-(k+1)}A^{k}\Big)B + D, }
\end{equation}
with region of convergence $|z|>\rho(A)$. Stability ($\rho(A)<1$) ensures that the boundary evaluation $H(e^{\jmath\omega})$ is well–defined and bounded for all $\omega\in[-\pi,\pi]$.

If $A$ is diagonalizable, $A=V\Lambda V^{-1}$ with $\Lambda=\mathrm{diag}(\lambda_1,\ldots,\lambda_N)$, then
\begin{equation}\label{eq:modal}
H(z) = \sum_{i=1}^{N} \frac{C v_i, w_i^\top B}{1-z^{-1}\lambda_i} + D,
\end{equation}
where $v_i$ and $w_i^\top$ are right/left eigenvectors ($V^{-1}=\begin{bmatrix}w_1^\top \\ \cdots \\ w_N^\top\end{bmatrix}$). Thus the poles of $H$ are precisely the eigenvalues of $A$; their radii $|\lambda_i|$ set decay rates, and their arguments $\arg(\lambda_i)$ set oscillation frequencies (discrete–time resonances).

Two norms are useful: the \emph{H$_\infty$} norm $\|H\|_{H_\infty}=\sup_{\omega}\sigma_{\max}(H(e^{\jmath\omega}))$ controls worst–case gain, while $\|H\|_{H_2}$ (finite for strictly proper $D=0$) relates to output energy under white excitation and equals $\mathrm{tr}(C W_c C^\top)$ where $W_c$ solves $W_c=AWA^\top+BB^\top$ \cite{kailath2000linear}.

\subsection{Impulse–response kernels, memory spectra, and Bode intuition}\label{sec:freq-impulse}

The time–domain input–output relation of \eqref{eq:lti} is a causal convolution
\begin{equation}\label{eq:impulse}
y_t = \sum_{k=0}^\infty h_k,u_{t-k},\qquad h_k := CA^{k}B \quad (k\ge 0), \quad h_{k<0}:=0.
\end{equation}
Hence the ESN’s linearized/lifted behavior is entirely encoded by the \emph{impulse kernel} $(h_k)_{k\ge0}$. Three basic consequences:

\paragraph{(i) Decay and “effective memory.”}
If $\rho(A)<1$, then $\|A^k\|\le c\,\rho(A)^k$ for some $c\ge1$. Therefore $\|h_k\|\le c\,\|C\|\,\|B\|\,\rho(A)^k$. The contribution of inputs older than $H$ time steps satisfies
$\sum_{k>H}\|h_k\| \lesssim \|C\|\,\|B\|\, \rho(A)^{H+1}/(1-\rho(A))$,
so the “$\varepsilon$–horizon” grows like $H_\varepsilon \simeq \log(\|C\|\|B\|/\varepsilon)/(1-\rho(A))$ (cf. Sec.~\ref{sec:props-fmp}). Near the edge of stability ($\rho(A)\uparrow 1$) the memory length diverges at rate $1/(1-\rho(A))$.

\paragraph{(ii) Modal oscillations and selectivity.}
Write $\lambda_i=r_i e^{\jmath\omega_i}$ with $0\!<\!r_i\!<\!1$. Modal contributions are
$h_k = \sum_i r_i^{\,k}\,e^{\jmath k\omega_i}\,(C v_i)(w_i^\top B),$
i.e., exponentially decaying sinusoids. Thus $H(e^{\jmath\omega})$ exhibits \emph{Bode} peaks near $\omega=\omega_i$; sharper peaks occur when $r_i$ is close to 1 or when the modal residue $(C v_i)(w_i^\top B)$ is large. Complex–conjugate pole pairs generate narrowband oscillatory memory; real poles near $+1$ generate broadband slow decay.

\paragraph{(iii) Power spectra under stochastic excitation.}
With zero–mean, wide–sense stationary input $u$ having spectral density $S_u(\omega)$, the output spectral density is
\begin{equation}\label{eq:wyner}
S_y(\omega) = H(e^{\jmath\omega})S_u(\omega)H(e^{\jmath\omega})^\ast,
\end{equation}
so $H$ shapes input spectra multiplicatively. For white input $S_u(\omega)=\sigma^2 I$, peaks of $S_y$ coincide with resonant modes of $A$. This ties kernel design to spectral selectivity in sequence tasks \cite{kailath2000linear}.

\paragraph{Bounds that connect time and frequency views.}
Let $W_c$ and $W_o$ be the reachability/observability Gramians of $(A,B)$ and $(A,C)$. Then
$\sum_{k\ge0}\mathrm{tr}(h_k h_k^\top) = \mathrm{tr}(C W_c C^\top) = \frac{1}{2\pi}\int_{-\pi}^{\pi}\mathrm{tr}(H(e^{\jmath\omega})H(e^{\jmath\omega})^\ast)\,d\omega,$
which equivalently views kernel energy as an average (over frequency) gain \cite{kailath2000linear}. Directions with large Gramian weight correspond to long-memory eigenmodes (eigenvalues near the unit circle).

\subsection{Relation to modern SSM kernels and when ESNs emulate them}\label{sec:freq-ssm-relation}

Modern “SSM layers” in deep sequence models (e.g., HiPPO and S4 families) implement long–range convolutions by parameterizing a stable state matrix with \emph{structured spectra} and learning the input/output couplings \cite{gu2020hippo,gu2022s4,gu2023mamba}. Their discrete- or continuous-time kernels take the canonical forms
\begin{equation}
k_k = C A^{k} B \quad\text{(DT)}, 
\qquad 
k(t) = C e^{A t} B \quad\text{(CT)},
\end{equation}
often with $A$ diagonal or normal in a known basis (e.g., diagonal–plus–low–rank (DPLR), Fourier/Legendre bases), enabling fast convolution via recurrences or FFTs. We now state precise conditions under which an ESN emulates these kernels.

\paragraph{(E1) Small–signal linearization with near–normal reservoir.}
Suppose the ESN operates in a regime where $J_\sigma(\xi)\approx \alpha I$ (e.g., symmetric slope about 0 for $\tanh$) so that the linearized $A\approx (1-\lambda)I+\lambda\alpha W$. If $W$ is chosen normal with prescribed eigenvalues (e.g., random unitary times diagonal radii) then $A$ inherits those eigenvectors and has poles $\lambda_i(A)=(1-\lambda)+\lambda\alpha\,\lambda_i(W)$. The resulting kernel $h_k=C A^{k}B$ is a mixture of decaying sinusoids whose rates and frequencies are controlled directly by the spectrum of $W$; this realizes the same family of exponentially decaying/oscillatory kernels parameterized in S4 via diagonal $A$ in a special basis \cite{gu2022s4}.

\emph{Design corollary.} To match a target kernel envelope with time constants $\{\tau_i\}$ and frequencies $\{\omega_i\}$, assign eigenvalues $\lambda_i(A)=e^{-1/\tau_i}\,e^{\jmath\omega_i}$ and choose $W\approx \frac{1}{\lambda\alpha}(\lambda_i(A)-(1-\lambda))$ along those modes; then set $B$ and $C$ (via $U$ and readout) to realize desired modal residues.

\paragraph{(E2) Lifted/Koopman LTI surrogates with DPLR structure.}
In a lifted ESN (Sec.~\ref{sec:lifted}), select a dictionary $\phi$ so that the identified $A_\phi$ is (approximately) diagonalizable with known basis and low–rank input/output couplings (e.g., $\phi$ includes identity and a small set of oscillatory features). If $A_\phi$ is diagonal and $B_\phi, C_\phi$ are low rank, then the kernel $h_k=C_\phi A_\phi^{k} B_\phi$ has the same DPLR structure exploited in S4 for $\tilde O(N)$ kernel generation and convolution \cite{gu2022s4}. The ESN thus “inherits” an SSM kernel through its lift, while its nonlinear core ensures the lift remains expressive on nonlinearly separable inputs.

\paragraph{(E3) Continuous–time equivalence and discretization.}
For CT parameterizations with $A_c$ Hurwitz (eigenvalues $\alpha_i+\jmath\omega_i$ with $\alpha_i<0$), the kernel $k(t)=C e^{A_c t}B=\sum_i e^{\alpha_i t}\,e^{\jmath\omega_i t}\,C v_i w_i^\top B$. Discretizing with step $\Delta t$ maps eigenvalues by $\lambda_i(A)=e^{(\alpha_i+\jmath\omega_i)\Delta t}$; therefore CT SSM kernels (e.g., HiPPO/S4) and DT ESN linearizations coincide after sampling. Tustin/bilinear discretization slightly shrinks radii (extra damping), which can be compensated in design (Sec.~\ref{sec:ct-disc}; \cite{franklin2015feedback}).

\paragraph{(E4) When the ESN does \emph{not} emulate structured SSM kernels.}
If $W$ is highly non-normal with widely spread pseudospectra, $A$ may produce long transients unrelated to its eigenvalues, leading to kernels that are not well captured by diagonal/DPLR parameterizations; conversely, steeply saturating $\sigma$ invalidates the $J_\sigma\approx \alpha I$ approximation, and the small–signal kernel changes with input amplitude. These regimes fall outside the clean equivalence with S4–style kernels and are better treated in the nonlinear time–domain (ISS) view.

\paragraph{Bode intuition and ESN hyperparameters.}
The leak $\lambda$ radially contracts/expands poles of $A$ (around $1$), hence sets the global memory scale; input scaling ($\|U\|$) and readout ($C$) scale modal residues; the reservoir spectrum (via $W$) sets pole radii and angles. Tuning $(\lambda,W,U)$ in ESNs is therefore equivalent, in the linearized/lifted sense, to shaping the SSM kernel’s passband, roll–off, and resonances—exactly the dials exposed by modern SSM layers \cite{gu2022s4,gu2020hippo}.

\section{Identification and Training via SSM Inference}\label{sec:id}

We cast ESN training and hyperparameter selection as inference in (linearized or lifted) state–space models. This yields denoised latent states, principled updates for noise covariances and dynamical hyperparameters, and spectral design knobs compatible with ESP/ISS.

\subsection{Teacher forcing as state estimation; KF/UKF/EKF perspectives}\label{sec:id-kf}

Consider the linearized/lifted surrogate
\begin{equation}\label{eq:id-lti}
x_{t+1}=A_t x_t + B_t u_t + w_t,\qquad y_t=C x_t + v_t,
\end{equation}
where $A_t\equiv A$ in the LTI case (Sec.~\ref{sec:small-signal}) or $A_t$ is frozen from a relinearization along a nominal trajectory (LTV). With Gaussian $w_t\!\sim\!\mathcal{N}(0,Q)$, $v_t\!\sim\!\mathcal{N}(0,R)$, the posterior over states is Gaussian and can be computed by Kalman filtering and Rauch–Tung–Striebel (RTS) smoothing \cite{rauch1965smoothing,sarkka2013bayesian,kailath2000linear}. Denote filter means/covariances by $(\mu_{t|t},P_{t|t})$ and one–step predictions by $(\mu_{t+1|t},P_{t+1|t})$. The forward recursions are
\begin{equation}
\mu_{t+1|t}=A_t\mu_{t|t}+B_t u_t,\quad P_{t+1|t}=A_t P_{t|t} A_t^\top+Q,
\end{equation}
\begin{equation}
K_t=P_{t|t-1}C^\top(C P_{t|t-1} C^\top+R)^{-1},\;\;
\mu_{t|t}=\mu_{t|t-1}+K_t(y_t-C\mu_{t|t-1}),\;\;
P_{t|t}=(I-K_t C)P_{t|t-1}.
\end{equation}

Backward smoothing refines these using $J_t=P_{t|t}A_t^\top P_{t+1|t}^{-1}$:
\begin{equation}
\mu_{t|T}=\mu_{t|t}+J_t(\mu_{t+1|T}-\mu_{t+1|t}),\quad
P_{t|T}=P_{t|t}+J_t(P_{t+1|T}-P_{t+1|t})J_t^\top,
\end{equation}
and the smoothed cross–covariance $P_{t,t+1|T}=J_t P_{t+1|T}$ (for LTI; the general formula includes an extra correction term) \cite{sarkka2013bayesian}.

In this lens, \emph{teacher forcing}—running the reservoir on the measured $u$ and using measured $y$ for supervision—becomes \emph{state estimation}: the smoothed means $\mu_{t|T} \approx \mathbb{E}[x_t\mid y_{1:T},u_{1:T}]$ are denoised “teacher–forced” states, with uncertainty $P_{t|T}$. For genuine nonlinear SSMs one uses EKF/RTS (linearize $f,g$ at $\mu$) or UKF/URTS (unscented transform) to approximate the same objects \cite{julier1997ukf,wan2000ukf,sarkka2013bayesian}. These posteriors are the sufficient statistics driving the EM updates below.

\subsection{EM for hyperparameters}\label{sec:id-em}

Let $\theta$ collect the parameters to estimate. In our setting $\theta$ can include (i) the discrete leak $\lambda$ or its CT time constant $\tau$ (Sec.~\ref{sec:ct-disc}), (ii) a \emph{spectral scaling} $\alpha$ multiplying the reservoir in small–signal form $A(\theta)\approx (1-\lambda)I+\lambda\,\alpha\,\bar W$ with $\bar W$ fixed (e.g., normalized $W$ or a Jacobian average), and (iii) noise covariances $Q,R$. With the linear–Gaussian surrogate \eqref{eq:id-lti}, maximum likelihood can be performed by EM \cite{shumway1982em,ghahramani1996lds,sarkka2013bayesian}.

\paragraph{E–step.}
Run a KF/RTS (or EKF/UKF smoother) under current $\theta^{(k)}$ to compute
\begin{equation}
\hat x_t:=\mu_{t|T},\quad \hat X_t:=P_{t|T}+\hat x_t \hat x_t^\top,\quad
\hat X_{t,t-1}:=P_{t,t-1|T}+\hat x_t \hat x_{t-1}^\top .
\end{equation}

Define the stacked regressor $Z_t:=[x_t^\top\; u_t^\top]^\top$ and its second moment $\hat Z_t Z_t^\top=\begin{bmatrix}\hat X_{t} & \hat x_t u_t^\top \\ u_t \hat x_t^\top & u_t u_t^\top\end{bmatrix}$.

\paragraph{M–step: $Q,R$.}
For fixed $A_t,B_t,C$ the covariance updates are closed–form:
\begin{align}
Q^{(k+1)} &= \frac{1}{T-1}\sum_{t=1}^{T-1}\Big(\hat X_{t+1}-A_t \hat X_{t}A_t^\top - B_t u_t \hat x_t^\top A_t^\top - A_t \hat x_t u_t^\top B_t^\top - B_t u_t u_t^\top B_t^\top\Big) \nonumber\\
&\quad + \frac{1}{T-1}\sum_{t=1}^{T-1}\Big( -\hat X_{t+1,t} A_t^\top - A_t \hat X_{t,t+1}\Big), \label{eq:Q-update}
\end{align}
\begin{align}
R^{(k+1)} &= \frac{1}{T}\sum_{t=1}^{T}\Big[(y_t-C\hat x_t)(y_t-C\hat x_t)^\top + C P_{t|T} C^\top\Big]. \label{eq:R-update}
\end{align}
The compact form is \(Q=\frac{1}{T-1}\sum \mathbb{E}[(x_{t+1}-A_t x_t-B_t u_t)(\cdot)^\top\mid y,u]\), \(R=\frac{1}{T}\sum \mathbb{E}[(y_t-Cx_t)(\cdot)^\top\mid y,u]\) \cite{ghahramani1996lds,sarkka2013bayesian}.

\paragraph{M–step: \(\lambda\), spectral scaling \(\alpha\).}
Treat \(A\) as \emph{structured linear} in \(\theta\): \(A(\theta)=\sum_{i=1}^{r}\theta_i M_i\) with basis \(M_1=I\), \(M_2=\bar W\) so that \(\theta_1=1-\lambda\), \(\theta_2=\lambda\alpha\). Maximizing the expected complete–data log–likelihood is equivalent to minimizing the quadratic prediction error
\begin{equation}
J(\theta,B)=\sum_{t=1}^{T-1}\mathbb{E}\!\left[\|x_{t+1}-A(\theta)x_t-B u_t\|_{Q^{-1}}^{2}\,\middle|\,y,u\right].
\end{equation}

Let $A_{\text{LS}}:=[\sum_t \hat x_{t+1}\hat x_t^\top - B \sum_t u_t \hat x_t^\top]\;[\sum_t \hat X_t]^{-1}$ be the (temporarily) unconstrained least–squares estimate of the state transition. Project $A_{\text{LS}}$ onto the span $\mathrm{span}\{M_1,M_2\}$ in Frobenius norm: write $M=[\mathrm{vec}(M_1)\;\mathrm{vec}(M_2)]$ and set
\begin{equation}\label{eq:theta-proj}
\hat\theta = (M^\top M)^{-1} M^\top\mathrm{vec}(A_{\text{LS}}),\qquad
\hat A = \hat\theta_1 I + \hat\theta_2 \bar W.
\end{equation}
Recover $\lambda,\alpha$ from $\hat\theta_1=1-\lambda$, $\hat\theta_2=\lambda\alpha$, and (optionally) project $(\lambda,\alpha)$ to $\lambda\!\in\!(0,1],\,\alpha\!>\!0$. To respect ESP/ISS, either enforce the conservative small–gain constraint $ (1-\lambda)+\lambda\,L_\sigma \|\alpha \bar W\|\!<\!1$ or the quadratic LMI certificate of Sec.~\ref{sec:props-contr} during the projection (nearest–feasible projection) \cite{khalil2002nonlinear,megretski1997iqc}.

Two refinements are standard. First, in an \emph{iterative local–linearization EM}, recompute $\bar W$ (e.g., as an average $J_\sigma(\hat\xi_t)W$) from the latest smoothed states and repeat. Second, in a lifted LTI model (Sec.~~\ref{sec:lifted}) the same EM applies with $A_\phi,B_\phi,C_\phi,Q,R$ unconstrained, after which $A_\phi$ can be reduced to a diagonal/DPLR form if desired (Sec.~~\ref{sec:freq}).

\subsection{Readout learning with state uncertainty: ridge/Bayesian forms}\label{sec:id-readout}

Given smoothed posteriors $\{\hat x_t,P_{t|T}\}$, the readout $C$ in $y_t=Cx_t+v_t$ is the maximizer of the conditional likelihood
\begin{equation}
\arg\min_C \sum_{t=1}^{T} \mathbb{E}\!\left[\|y_t-C x_t\|_{R^{-1}}^{2}\,\middle|\,y,u\right].
\end{equation}

This yields the closed form
\begin{equation}\label{eq:C-ml}
\hat C = \Big(\sum_{t} y_t \hat x_t^\top\Big)\Big(\sum_{t} (\hat X_t)\Big)^{-1},
\qquad \hat X_t:=P_{t|T}+\hat x_t\hat x_t^\top.
\end{equation}
Adding Tikhonov (ridge) regularization $ \tfrac{\lambda_r}{2}\|C\|_F^2$ modifies the denominator to $\sum_t \hat X_t + \lambda_r I$.

A fully Bayesian treatment places a Gaussian prior $\mathrm{vec}(C)\sim\mathcal{N}(0,\tau^{-1}I)$. Approximating the latent state by its posterior (Laplace/plug–in), the posterior over $\mathrm{vec}(C)$ is Gaussian with precision
\begin{equation}\label{eq:C-bayes}
\Lambda_C = \tau I + \sum_{t} ( \hat X_t \otimes R^{-1} ),
\qquad
\mathrm{vec}(\hat C)=\Lambda_C^{-1}\sum_t ( \hat x_t \otimes R^{-1}),y_t ,
\end{equation}
where $\otimes$ is the Kronecker product \cite{bishop2006prml}. This quantifies readout uncertainty inherited from state uncertainty. In the lifted LTI case one can similarly update $D$ (if present) and treat multi–output couplings.

\subsection{Hybrid subspace identification to shape reservoir spectra under ESP}\label{sec:id-subspace}

Subspace identification (SSI) offers nonparametric estimates of $A,B,C$ from input–output trajectories via Hankel matrices and orthogonal projections (MOESP/N4SID) \cite{vanoverschee1996subspace,ljung1999system}. In the ESN–as–SSM view, SSI can \emph{shape} the reservoir spectrum in a data–driven but stability–aware manner.

The workflow is conceptual. First, compute an SSI estimate $(A_{\text{ssi}},B_{\text{ssi}},C_{\text{ssi}})$ for a chosen model order on persistently exciting data. This gives a target pole set and residue structure that match the task’s memory spectrum (Sec.~\ref{sec:freq}). Second, \emph{project} this target onto the ESN’s small–signal form. With $A(\theta)=\theta_1 I+\theta_2 \bar W$ as above, solve
\begin{equation}\label{eq:proj-ssi}
\min_{\theta} |A(\theta)-A_{\text{ssi}}|_F^2 \quad \text{s.t.}\quad \text{ESP/ISS certificate (e.g., \eqref{eq:C1} or \eqref{eq:C2LMI})},
\end{equation}
using the linear projection \eqref{eq:theta-proj} followed by a nearest–feasible projection onto the contractive set. If one is allowed to \emph{design} $W$, choose $W$ normal with eigenvalues that pull the poles of $A(\theta)$ toward those of $A_{\text{ssi}}$ while enforcing $(1-\lambda)+\lambda L_\sigma \|\alpha W\|<1$ (or the LMI). This preserves ESP/ISS and imports SSI’s spectral profile.

When using lifted models, apply SSI on the lifted state $z_t$ (estimated via EDMD features) to obtain $A_\phi$ with a diagonal/DPLR structure; this immediately delivers S4–like kernels (Sec.~\ref{sec:freq}) while the original ESN nonlinearity governs feature construction. In all cases, persistent excitation of inputs and well–conditioned Gramians are essential for identifiability and low variance \cite{vanoverschee1996subspace,ljung1999system}.

\paragraph{Remarks.}
(i) EM and SSI can be combined: use SSI to initialize $(A,B,C)$, then refine $Q,R$ and the structured $\theta$ by EM with a contraction constraint. (ii) Enforcing ESP/ISS during identification is not merely cosmetic: it regularizes ill–posed long–memory fits and mitigates drift, echoing the small–gain/LMI design in Sec.~~\ref{sec:props-contr}. (iii) In CT formulations, the same pipeline applies with $A_c$ Hurwitz, then discretized (Sec.~~\ref{sec:ct-disc}).

\section{Design Recipes and Practical Guidelines}\label{sec:design}

We synthesize system–theoretic insights into practical knobs for ESNs viewed as SSMs. Let $A$ denote a linearized/lifted state matrix (Sec.~~\ref{sec:small-signal}, \ref{sec:lifted}), with contraction factor $\kappa:=\|A\|$ (or a certified $\kappa<1$ in a weighted norm via Sec.~~\ref{sec:props-contr}). Unless noted, $\sigma=\tanh$ so $L_\sigma\le 1$.

\subsection{Choosing leak, spectral radius, input scaling: rules of thumb from poles \& kernels}\label{sec:design-leak}

\paragraph{Target memory $\Rightarrow$ pole radius.}
Pick an \emph{effective horizon} $H$ in steps at tolerance $\varepsilon$. For a stable LTI surrogate, contributions older than $H$ scale like $\rho(A)^H$. A convenient target is the half–life $H_{1/2}=\ln 2/(-\ln \rho(A))$. Equivalently,
\begin{equation}
\label{eq:target-r}
\rho(A)\approx r_\star := 2^{-1/H_{1/2}}\approx \exp\big(-1/H\big),
\end{equation}
(cf. the $\varepsilon$–horizon estimate in \eqref{eq:Heps}). Choose $r_\star$ small for short memory; push $r_\star\uparrow 1$ for long memory (noting the $1/(1-r_\star)$ sensitivity trade–off).

\paragraph{Mapping $(\lambda,W)\mapsto \rho(A)$.}
Around typical operating slopes $s:=\mathbb{E}[\sigma'(\xi)]\in(0,L_\sigma]$, the linearization reads
\begin{equation}
A \;\approx\; (1-\lambda)I \;+\; \lambda\,s\,W.
\end{equation}

If $W$ is normal with $\rho(W)=\gamma$, then $\rho(A)\approx (1-\lambda)+\lambda s\gamma$. To hit $r_\star$ for a chosen $\lambda$,
\begin{equation}
\label{eq:gamma-choice}
\gamma \approx \frac{r_\star-(1-\lambda)}{\lambda s};\;\ \text{(clip to }\gamma\in(0,1/L_\sigma)\text{ for ESP margin)}.
\end{equation}
When $s$ is uncertain, design with $s=L_\sigma$ to preserve a guaranteed ESP margin; in practice $s\in[0.5,1)$ for $\tanh$ if preactivations are near zero–mean with unit scale, yielding less conservative $\gamma$.

\paragraph{Leak as a time–constant.}
In CT form $\tau\dot x=-x+\sigma(\cdot)$ and explicit Euler with step $\Delta t$ gives $\lambda=\Delta t/\tau$ (Sec.~\ref{sec:ct-disc}). Thus $\tau$ sets a global decay scale. For multi–rate memory, use \emph{block leaks} $\lambda=\mathrm{blkdiag}(\lambda_1,\ldots,\lambda_G)$ (or stacked reservoirs), covering a log–spaced set of time constants $\tau_g$ whose induced $\rho(A_g)$ tile $(r_{\min},r_{\max})$.

\paragraph{Spectrum shaping.}
Prefer $W$ \emph{normal} (e.g., unitary times diagonal radii) so $A$ inherits well–behaved modes; place a fraction of poles as real radii near $+1$ (slow trends) and the rest as complex–conjugate pairs with $|\lambda_i|=r_\star$ and $\arg(\lambda_i)$ tiling $[0,\pi]$ (oscillatory memory). A \emph{log–uniform} spread of radii $|\lambda_i|\in[r_{\min},r_{\max})$ approximates a $1/f$–like bank of time scales.

\paragraph{Guaranteed ESP margin.}
A conservative global certificate (Sec.~\ref{sec:prelim-esn}, \ref{sec:props-contr}) is
\begin{equation}
(1-\lambda)+\lambda\,L_\sigma\,\|W\| \;<\; 1 \quad\Longleftarrow\quad \|W\|\;<\;1/L_\sigma,
\end{equation}
independent of $\lambda$. In practice enforce $\|W\|\le \eta/L_\sigma$ with $\eta\in(0,1)$ to retain margin under input–dependent slope variation; then tune $\lambda$ and the \emph{modal placement} within that budget to reach $r_\star$. For tighter, less conservative guarantees, certify contraction in a weighted norm via the LMI in \eqref{eq:C2LMI}.

\paragraph{Input scaling $U$: keep $\xi=Wx+Uu+b$ in the linear–slope region.}
Favour regimes where $|\xi_i|\lesssim 2$ so $\sigma'(\xi)$ stays large (for $\tanh$). Whiten inputs (pre–scale $u$ so $\operatorname{Cov}(u)\approx I$), then choose $U$ with per–row norm $\|U_{i:}\|_2\approx \upsilon$ producing $\operatorname{Var}(\xi_i)\approx \upsilon^2$ near $1$. A simple default is $\|U\|_F^2\approx n$ (unit variance per neuron when $u$ is white). If the reservoir already yields $\operatorname{Var}(Wx)$ near unity, reduce $\upsilon$ to avoid saturation; if $Wx$ is small (strong leak), increase $\upsilon$ to preserve sensitivity. The input Lipschitz $L_u=\lambda\|U\|L_\sigma$ (Sec.~\ref{sec:prelim-esn}) quantifies the gain/noise trade–off.

\subsection{Noise modeling as regularization; robustness to drift \& shift}\label{sec:design-noise}

\paragraph{Process/measurement noise as \emph{state} and \emph{output} regularizers.}
In a linearized/lifted SSM with Gaussian $w_t\sim\mathcal{N}(0,Q)$, $v_t\sim\mathcal{N}(0,R)$, the maximum–likelihood state estimate minimizes
\begin{equation}
\sum_{t=1}^{T-1}\|x_{t+1}-Ax_t-Bu_t\|_{Q^{-1}}^2 + \sum_{t=1}^{T}\|y_t-Cx_t\|_{R^{-1}}^2,
\end{equation}
so $Q$ penalizes fast variations of the latent dynamics (a \emph{temporal} Tikhonov term), while $R$ penalizes readout misfit (a \emph{measurement} weighting). Larger $Q$ induces smoother, more robust hidden states; larger $R$ downweights noisy outputs and curbs overfitting of $C$ (Sec.~\ref{sec:id-kf}, \ref{sec:id-readout}) \cite{sarkka2013bayesian,kailath2000linear}.

\paragraph{ISS robustness bound.}
With contraction factor $\kappa<1$, additive model error/disturbance $d_t$ (e.g., RF approximation error, shifts) produces a bounded state deviation
\begin{equation}
\|x_t-\bar x_t\|\;\le\;\kappa^t\|x_0-\bar x_0\|+\sum_{k=0}^{t-1}\kappa^{t-1-k}\|d_k\|
\;\le\; \frac{\sup_k \|d_k\|}{1-\kappa},
\end{equation}
so designing for a margin $1-\kappa$ trades longer memory for bounded sensitivity (Sec.~\ref{sec:props-fmp}) \cite{sontag2008iss,angeli2002lyapunov}. This is the core robustness lever under drift/shift: keep $\kappa$ comfortably below $1$ when reliability trumps horizon.

\paragraph{Distribution shift: forgetting and covariance scheduling.}
Under nonstationarity, use \emph{exponential forgetting} in readout estimation (recursive ridge/least squares with factor $\beta\in(0,1)$) to bias toward recent regimes; in SSM inference, increase $Q$ (or reduce $\beta$) when mismatch grows, and decrease $R$ when outputs are trusted—both are principled levers in KF/RTS that correspond to adaptive regularization \cite{sarkka2013bayesian}. For shifts that alter the signal’s passband, re–place poles (Sec.~\ref{sec:design-leak}) or switch among predesigned spectral tiles (block leaks) rather than pushing a single $\kappa$ to the edge.

\paragraph{Sector/LMI robustness to slope variation.}
If inputs can drive $\sigma'$ anywhere in $[0,L_\sigma]$, certify absolute stability with the LMI in \eqref{eq:C2LMI} (discrete circle–criterion). This gives a \emph{uniform} ESP certificate that tolerates slope fluctuations and small modeling errors without re–tuning $\lambda$ or $\|W\|$ \cite{khalil2002nonlinear,megretski1997iqc}.

\subsection{Computational complexity, batching, and long–sequence throughput}\label{sec:design-compute}

\paragraph{Per–step update.}
A dense reservoir update costs $O(n^2)$ flops for $Wx_t$ plus $O(nm)$ for $Uu_t$ and $O(n)$ for $\sigma(\cdot)$. With $k$ nonzeros per row (sparsity), the cost is $O(kn)$. Orthogonal/FFT–like $W$ (Hadamard/DFT with diagonal scalings) yields $O(n\log n)$ multiplies while preserving normality and good spectra.

\paragraph{Readout estimation.}
Closed–form ridge on $T$ steps forms the Gram $\sum_t x_t x_t^\top$ in $O(Tn^2)$ and inverts an $n\times n$ matrix in $O(n^3)$. For long sequences, maintain running sums (streaming ridge) and solve once; for very large $n$, use conjugate gradients on the normal equations with the Gram as a linear operator. With state uncertainty (Sec.~\ref{sec:id-readout}), replace $x_t x_t^\top$ by $\hat X_t=P_{t|T}+\hat x_t\hat x_t^\top$.

\paragraph{Kalman smoothing.}
Dense KF/RTS is $O(n^3)$ per time step due to Riccati updates. Structure drops this: diagonal/banded $A$ yields $O(n)\!/O(nb^2)$; normal $A=Q\Lambda Q^\ast$ with isotropic $Q$ allows fast transforms in prediction/innovation covariances. In lifted SSMs with diagonal/DPLR $A$ (Sec.~\ref{sec:lifted}, \ref{sec:freq}), convolutions can be executed in $O(N\log N)$ via FFT–style kernels.

\paragraph{Batching and long sequences.}
For $B$ sequences of length $L$, teacher forcing is embarrassingly parallel across sequences. In memory–bound settings, use \emph{chunked} unrolling: process windows of length $L_w$ with overlap $H$ (the effective horizon), carrying only the final states between chunks—errors beyond $H$ are negligible by design (cf. \eqref{eq:Heps}). Keep computations in float64 when $\rho(A)\approx 1$ to avoid accumulation error; renormalize states intermittently if using non–normal $W$.

\paragraph{When to linearize/lift for speed.}
If the downstream pipeline only needs linear responses (e.g., fixed kernels), freeze a lifted LTI surrogate with certified stability and run $\,y_t=\sum_{k\ge0}h_k u_{t-k}$ via FFTs (Sec.~\ref{sec:freq}). Use the full nonlinear ESN only when input–dependent gating is essential; otherwise linearized kernels offer substantial throughput gains with predictable memory spectra.

\medskip
Taken together: (i) pick a memory target and place poles (via $\lambda$, $\rho(W)$, and modal geometry) to hit $r_\star$; (ii) keep preactivations in the linear–slope regime by whitening inputs and scaling $U$; (iii) retain an ESP margin (global Lipschitz or LMI certificate) for robustness; (iv) treat $Q,R$ as principled regularizers against drift and noise; and (v) exploit structure (normal/sparse $W$, diagonal/DPLR lifts) to keep cost linear or near–linear in $n$ and sequence length.

\section{Extensions}\label{sec:ext}

\subsection{Deep/stacked ESNs as block SSMs; skip connections and multi–rate leaks}\label{sec:ext-deep}

Consider $L$ leaky ESN layers with intra–layer recurrent weights $W_\ell$, inter–layer feedforward $V_\ell$ (from layer $\ell{-}1$ to $\ell$), input matrices $U_\ell$, and leaks $\lambda_\ell\in(0,1]$:
\begin{equation}
\label{eq:deep-esn}
\begin{aligned}
x^{(1)}_{t+1} &= (1-\lambda_1)\,x^{(1)}_{t}
               + \lambda_1\,\sigma\!\big(W_1 x^{(1)}_{t} + U_1 u_t + b_1\big),\\
x^{(\ell)}_{t+1} &= (1-\lambda_\ell)\,x^{(\ell)}_{t}
               + \lambda_\ell\,\sigma\!\big(W_\ell x^{(\ell)}_{t}
               + V_\ell x^{(\ell-1)}_{t} + U_\ell u_t + b_\ell\big),
               \qquad \ell=2,\dots,L,\\
y_t &= C \begin{bmatrix} x^{(1)}_{t} \\ \vdots \\ x^{(L)}_{t} \end{bmatrix} + d .
\end{aligned}
\end{equation}
with optional \emph{skip connections} (residuals) from earlier layers to the readout. Linearizing around an operating trajectory produces a block lower–triangular SSM
\begin{equation}
\delta X_{t+1}=\underbrace{\begin{bmatrix}
A_{11} & 0      & \cdots & 0\\
A_{21} & A_{22} & \cdots & 0\\
\vdots & \ddots & \ddots & \vdots\\
A_{L1} & \cdots & A_{L,L-1} & A_{LL}
\end{bmatrix}}_{=:A_{\mathrm{blk}}}\delta X_t
+
\underbrace{\begin{bmatrix} B_1 \\ \vdots \\ B_L\end{bmatrix}}_{=:B_{\mathrm{blk}}}\delta u_t,\qquad 
\delta y_t = C\,\delta X_t,
\end{equation}
where $A_{\ell\ell}=(1-\lambda_\ell)I+\lambda_\ell J_\sigma(\xi_\ell)W_\ell$ and $A_{\ell,\ell-1}=\lambda_\ell J_\sigma(\xi_\ell)V_\ell$. Because $A_{\mathrm{blk}}$ is block triangular,
\begin{equation}
\rho(A_{\mathrm{blk}})=\max_{\ell}\rho(A_{\ell\ell}).
\end{equation}
Hence, if each layer satisfies the single–layer contraction (e.g., $(1-\lambda_\ell)+\lambda_\ell L_\sigma\|W_\ell\|<1$), then the deep stack is (globally) exponentially stable in the linearized sense; with slope–restricted $\sigma$, the weighted–norm LMI (circle–criterion/IQC) yields a \emph{uniform} ESP certificate for the full deep ESN (Sec.~\ref{sec:props-contr}; \cite{khalil2002nonlinear,megretski1997iqc}).

\paragraph{Multi–rate leaks.}
Choosing $\{\lambda_\ell\}$ on a log–spaced grid of time constants $\tau_\ell=\Delta t/\lambda_\ell$ tiles a bank of memory scales—equivalently, $A_{\ell\ell}$ places pole radii $|\lambda_i(A_{\ell\ell})|$ across $[r_{\min},r_{\max})$. Residual/skip readouts improve observability by exposing both short– and long–memory layers to the linear head. See also deep/stacked RC \cite{gallicchio2017deepesn,lukosevicius2012practical}.

\subsection{Bilinear/multiplicative SSM formulations for input–gated reservoirs}\label{sec:ext-bilinear}

Input–dependent \emph{gating} can be expressed bilinearly. Two equivalent views:

\paragraph{(i) Bilinear SSM.}
\begin{equation}\label{eq:bilinear-ext}
x_{t+1} = A x_t + B u_t + \sum_{i=1}^{m} u_{t,i}B_i x_t + w_t,\qquad y_t=Cx_t+Du_t+v_t,
\end{equation}
with $B_i$ capturing multiplicative modulation by input channel $i$ \cite{kailath2000linear}. This arises from a first–order expansion of $J_\sigma(\xi)$ around a nominal $\bar \xi$ (Sec.~\ref{sec:small-signal}: “bilinearized surrogates”).

\paragraph{(ii) Gated ESN map.}
\begin{equation}\label{eq:gated-esn}
x_{t+1}
= (1-\lambda)\,x_t
+ \lambda\,\sigma\!\Big(
  W x_t
  + \underbrace{\,\underbrace{\Gamma(u_t)}_{\text{diag or low-rank}}\;\tilde W x_t\,}_{\text{multiplicative gate}}
  + U u_t + b
\Big).
\end{equation}
with $\Gamma(u_t)$ diagonal or low rank (e.g., $\Gamma(u_t)=\mathrm{diag}(G u_t)$). Under bounded $u_t$ and slope–restricted $\sigma$, the closed loop admits a contraction certificate if
\begin{equation}
(1-\lambda)+\lambda L_\sigma\big(\|W\|+\|\Gamma(u)\|\|\tilde W\|\big) < 1 \quad \text{uniformly over } u\in\mathcal{U}_M,
\end{equation}
or via the quadratic LMI of Sec.~\ref{sec:props-contr} (replace $B_z$ and $G_x$ accordingly). Bilinear/gated SSMs retain linear readouts and enable selective memory injection without sacrificing ESP \cite{khalil2002nonlinear,megretski1997iqc}.

\subsection{Spatiotemporal inputs: convolutional reservoirs as local SSMs on grids}\label{sec:ext-conv}

Let images $u_t\in\mathbb{R}^{H\times W\times c}$, states $x_t\in\mathbb{R}^{H\times W\times n}$. A \emph{local SSM} with periodic boundaries is
\begin{equation}\label{eq:conv-ssm}
x_{t+1}=A_\ast \ast x_t + B_\ast \ast u_t + w_t,\qquad y_t=C_\ast \ast x_t + v_t,
\end{equation}
where $\ast$ denotes spatial convolution and $A_\ast$ is a bank of $n{\times}n$ kernels (e.g., $3{\times}3$ or $5{\times}5$). Vectorizing the field reveals a block–circulant–with–circulant–blocks (BCCB) matrix $\mathcal{A}$ whose eigenvectors are 2D Fourier modes; the \emph{symbol} $\widehat A(\omega_x,\omega_y)\in\mathbb{R}^{n\times n}$ is the DFT of the kernels. Stability and memory follow \emph{frequencywise}:
\begin{equation}
\rho(\mathcal{A}) \;=\; \sup_{(\omega_x,\omega_y)} \rho\!\big(\widehat A(\omega_x,\omega_y)\big)\;<\;1
\quad\Longleftrightarrow\quad \text{DT exponential stability / BIBO}. 
\end{equation}
Thus design can place per–frequency poles (lowpass/ bandpass/ directional) by shaping $\widehat A$, exactly paralleling CNN intuition with system guarantees \cite{kailath2000linear,gray2006toeplitz}.

A \emph{convolutional ESN} instantiates \eqref{eq:conv-ssm} by replacing $A_\ast \ast x_t$ with $(1-\lambda) x_t + \lambda \sigma((W_\ast \ast x_t)+(U_\ast \ast u_t)+b)$, where $W_\ast$ and $U_\ast$ are random fixed kernels (shared spatially). The small–signal $A_\ast(\xi)=(1-\lambda)I+\lambda J_\sigma(\xi) W_\ast$ yields a BCCB $A$; the same frequencywise radius test applies. On graphs or irregular grids, replace convolutions by polynomials of the (normalized) Laplacian $L$: $A\approx (1-\lambda)I+\lambda\,s\,\sum_{k=0}^{K}\alpha_k T_k(\tilde L)$, with Chebyshev polynomials $T_k$ diagonalized by the graph Fourier basis; stability holds if the polynomial stays inside the unit disk on the Laplacian spectrum \cite{kailath2000linear}.

\subsection{Probabilistic ESNs: process/measurement noise, smoothing, and uncertainty}\label{sec:ext-prob}

Adopt the linearized/lifted model with Gaussian noise:
\begin{equation}
x_{t+1}=Ax_t+Bu_t+w_t,\quad w_t\sim\mathcal{N}(0,Q),\qquad y_t=Cx_t+v_t,\quad v_t\sim\mathcal{N}(0,R).
\end{equation}

Kalman filtering/smoothing yields posteriors $p(x_t\mid y_{1:T},u_{1:T})=\mathcal{N}(\mu_{t|T},P_{t|T})$ (Sec.~\ref{sec:id-kf}; \cite{sarkka2013bayesian,kailath2000linear}). The \emph{predictive} distribution at horizon $h$ is
\begin{equation}
p(y_{t+h}\mid y_{1:t},u_{1:t+h})=\mathcal{N}\!\Big(C A^{h}\mu_{t|t}+ \sum_{j=0}^{h-1} C A^{j} B u_{t+h-1-j}, \;\; C \Sigma_h C^\top + R\Big),
\end{equation}
with $\Sigma_h = A^h P_{t|t} (A^\top)^h + \sum_{j=0}^{h-1} A^{j} Q (A^\top)^{j}$. Credible bands follow immediately. In the nonlinear ESN, EKF/UKF provide Gaussian approximations to the same objects. Bayesian readouts (ridge priors) propagate latent uncertainty into output uncertainty (Sec.~~\ref{sec:id-readout}; \cite{bishop2006prml}). These probabilistic views formalize ESN robustness to noise/drift and support principled model selection through marginal likelihood (EM; Sec.~~\ref{sec:id-em}).

\medskip

\section{Related Work}\label{sec:related}

\subsection{Reservoir computing \& fading–memory theory}\label{sec:related-rc}

Echo State Networks and Liquid State Machines introduced the RC paradigm of fixed recurrent cores with trained linear readouts \cite{jaeger2001echo,maass2002real}. Foundational analyses connect RC to causal fading–memory filters \cite{BoydChua1985}, establish universality and rates for ESN–like architectures under Lipschitz/mixing assumptions \cite{grigoryeva2018echo,gonon2020riskbounds}, and characterize the ESP via contraction and input–to–state stability (ISS) \cite{manjunath2013}. Practical surveys and tutorials cover design choices and training (ridge, teacher forcing) \cite{lukosevicius2012practical}. Our contribution recasts these results in standard SSM language, enabling direct use of ISS, Gramians, and frequency–domain tools.

\subsection{System identification and SSMs (classical to modern)}\label{sec:related-id}

The classical LTI/SSM canon (transfer functions, BIBO, controllability/observability, Gramians) is well established \cite{kailath2000linear}. Identification methods include subspace identification (MOESP/N4SID) and likelihood/EM for linear–Gaussian models \cite{vanoverschee1996subspace,ljung1999system,ghahramani1996lds,shumway1982em}, with Kalman filtering/smoothing as the core inference engine \cite{sarkka2013bayesian,rauch1965smoothing}. Our SSM view of ESNs imports these tools: state denoising via KF/RTS, EM updates for $Q,R$ and spectral scaling under ESP constraints, and SSI–based spectral shaping. For nonlinear systems, contraction/ISS and circle–criterion/IQC LMIs provide robust, slope–restricted stability certificates \cite{khalil2002nonlinear,megretski1997iqc}.

\subsection{Connections to neural ODE/CDE and kernel/state–space Transformers}\label{sec:related-ssmdeep}

Neural ODEs and CDEs parameterize flows $\dot x=f_\theta(x,t)$ and controlled dynamics $dx=f_\theta(x,t)\,dt+g_\theta(x,t)\,dU_t$, connecting deep models to continuous–time dynamical systems and rough paths \cite{chen2018neuralode,kidger2020neuralcde}. Parallelly, modern sequence models based on \emph{structured state space} layers (HiPPO, S4, and successors) learn long–range convolutional kernels $k_k=CA^{k}B$ (or $k(t)=Ce^{At}B$) with spectra designed for stability and efficiency \cite{gu2020hippo,gu2022s4,gu2023mamba}. Our analysis shows that ESNs, under linearization or lifting, yield precisely such kernels: poles of the small–signal $A$ set decay/oscillation; residues determine passbands; leaks control global time–constants. The SSM lens therefore provides a rigorous bridge between classical RC and contemporary kernel/SSM architectures, while ESP/ISS–based certificates offer principled stability beyond heuristic spectral–radius tuning.

\section{Discussion and Limitations}\label{sec:disc}

\subsection{When ESN-as-SSM helps—and when it does not}\label{sec:disc-when}
Casting ESNs as SSMs is most advantageous when the driven dynamics admit a uniform contraction (or a verifiable weighted contraction) on a forward–invariant set and when the task can be characterized by stable, exponentially decaying (possibly oscillatory) memory kernels. In these regimes, poles and residues give interpretable handles on horizon, selectivity, and gain; ISS certificates translate directly into Echo State guarantees; and identification tools (Kalman smoothing, EM, and subspace methods) become well posed, providing denoised latent states and principled hyperparameter updates (Secs.~~\ref{sec:props}–\ref{sec:id}). The frequency–domain view then supplies Bode–style intuition for passbands and roll–off (Sec.~~\ref{sec:freq}), while structured designs—normal or diagonal–plus–low–rank state matrices, block leaks, and convolutional/graph local SSMs—deliver predictable behavior with efficient implementations (Secs.~\ref{sec:design}, \ref{sec:ext}).

The reduction is less faithful when the update departs strongly from the slope–restricted regime or when input–dependent switching dominates. Highly non–Lipschitz maps (e.g., polynomial activations without domain restriction), hard resets or discontinuities, and aggressive multiplicative gating can break a uniform contraction and invalidate small–signal surrogates except on tiny tubes. Severe saturation is also problematic: if the preactivation spends long intervals where $\sigma'(\xi)\!\approx\!0$, then the effective Jacobian becomes state–dependent in a way that undermines any frozen $A$, and the lifted LTI approximation must be reidentified continually (Sec.~\ref{sec:small-signal}). Strong non–normal reservoirs create large transient growth unrelated to eigenvalues; linearized pole placement then mispredicts short–timescale amplification even if $\rho(A)\!<\!1$. Finally, tasks that require algebraically long memory, noncausal context, or exact time–delay lines are not well captured by exponentially decaying kernels; in such cases (e.g., nearly lossless propagation or delay–dominant physics), the SSM kernel family needs augmentation with near–unit–modulus modes or explicit delay embeddings.

\subsection{Identifiability caveats, data requirements, and brittleness to mis–specification}\label{sec:disc-ident}
State–space parametrizations are identifiable only up to similarity transforms; consequently, linearized or lifted models admit multiple equivalent representations with indistinguishable input–output behavior. In lifted spaces, nonuniqueness is amplified: different dictionaries $\phi$ (EDMD, random features, Carleman truncations) induce distinct coordinates $z$ and hence distinct $(A_\phi,B_\phi,C_\phi)$ that approximate the same nonlinear filter on a compact set. Without further structure, diagonal or DPLR reductions are model choices rather than truths. This places a premium on explicit structural priors (normality, bandedness, diagonal blocks) and on stability certificates that are invariant to coordinate changes (ISS/LMI conditions in Sec.~\ref{sec:props-contr}).

Reliable identification also demands sufficient excitation and window length. Subspace methods require persistently exciting inputs of order equal to the lifted dimension and observation windows covering the effective horizon; EM needs adequate signal–to–noise and well–conditioned Gramians to avoid degenerate $Q$/$R$ estimates. In practice, confounding between leak $\lambda$ and spectral scaling of $W$ is common: multiple $(\lambda,\alpha)$ pairs can realize nearly the same pole radii in $A$. Regularization or constraints (e.g., fixing $\|W\|$ and estimating $\lambda$) mitigate this ambiguity, but the resulting point estimates should be interpreted as one of several observationally equivalent parametrizations.

Model mis–specification introduces brittleness that the SSM lens makes explicit. Gaussian, time–invariant $Q$/$R$ are convenient but often unrealistic; colored disturbances, heavy tails, and nonstationary noise alter Kalman gains and can bias EM. Piecewise–stationary data induce LTV linearizations whose frozen $A_t$ change faster than identification can track; if relinearization is too infrequent, the kernel view drifts from the true behavior. Approximate closure errors in lifts (finite dictionaries) act as persistent disturbances; their accumulation scales as $O(\varepsilon/(1-\kappa))$ and can dominate when the design is pushed to the edge of stability. These effects highlight the value of maintaining a stability margin (not just $\rho(A)\!<\!1$ but a certified contraction) and of treating $Q$ as a tunable regularizer rather than a physical covariance (Secs.~\ref{sec:design-noise}, \ref{sec:id-em}).

\subsection{Opportunities for structure}\label{sec:disc-struct}
The SSM framing surfaces principled opportunities for structure that trade a small amount of generality for strong guarantees and speed. Normal or near–normal $A$ provides transparent pole geometry and well–behaved transient growth; diagonal–plus–low–rank forms preserve expressivity while enabling fast kernel application and stable identification. Banded or Toeplitz–like $A$ captures local couplings and admits $O(N)$–$O(N\log N)$ algorithms; convolutional and graph–polynomial parametrizations extend this to images and networks with per–frequency stability checks (Sec.~~\ref{sec:ext-conv}). Multi–rate leaks and block–triangular stacks orchestrate time–scale tiling without losing global ESP because the block spectral radius reduces to the worst layer (Sec.~~\ref{sec:ext-deep}). Finally, sector/IQC LMIs furnish coordinate–free stability certificates that tolerate slope variation, allowing one to dial memory close to the boundary while maintaining robustness (Sec.~~\ref{sec:props-contr}). Each of these structures integrates naturally with the probabilistic and identification pipelines of Sec.~~\ref{sec:id}, encouraging stable, interpretable, and efficient ESN designs.

\section{Conclusion}\label{sec:conclusion}
We have argued that Echo State Networks sit squarely inside the state–space modeling landscape. Written as a nonlinear SSM, a leaky ESN inherits ISS as the proper translation of the Echo State Property; linearizations and lifts expose poles, residues, and impulse kernels that determine memory and selectivity; and standard inference tools—Kalman smoothing, EM, and subspace identification—supply denoised states, principled hyperparameter updates, and spectral shaping. This unified view turns familiar ESN heuristics into certified design rules: leaks become time constants; spectral scaling becomes pole placement; kernel decay and oscillation become Bode–style dials; and stability moves from spectral–radius folklore to contraction certificates and LMIs. At the same time, the framework delineates its limits—non–Lipschitz regimes, heavy switching, non–normal transients, and long–delay tasks—and suggests remedies rooted in structure, margin, and probabilistic regularization.

Beyond clarifying practice, the perspective opens concrete research directions: contraction–metric learning and data–driven certificates for ESNs; principled dictionary selection for lifts that balance closure with stability; structured reservoirs (normal, DPLR, banded) with fast kernels and end–to–end guarantees; multi–rate and convolutional ESNs for spatiotemporal domains; and probabilistic ESNs with calibrated uncertainty under drift. By bridging classical RC and modern SSM layers, the analysis provides a common language in which stability, identifiability, and efficiency are first–class citizens—an inviting basis for the next generation of reservoir designs.

\bibliographystyle{abbrv}
\bibliography{ref}

\end{document}